\documentclass{article}

    \usepackage[final]{neurips_2025}

\usepackage[utf8]{inputenc} 
\usepackage[T1]{fontenc}    
\usepackage{hyperref}       
\usepackage{url}            
\usepackage{booktabs}       
\usepackage{amsfonts}       
\usepackage{nicefrac}       
\usepackage{microtype}      
\usepackage{xcolor}         
\usepackage{graphicx} 
\usepackage{amsmath}
\usepackage{multirow}
\usepackage{colortbl}
\usepackage{pifont}
\usepackage{subfig}

\usepackage{xspace}
\usepackage{wrapfig}
\usepackage{array}
\usepackage[table]{xcolor}
\usepackage{pifont}
\usepackage{wrapfig} 

\newcommand{\pub}[1]{\color{gray}{\scriptsize{[{#1}]}}}
\makeatletter
\newcommand{\thickhline}{
    \noalign {\ifnum 0=`}\fi \hrule height 1pt
    \futurelet \reserved@a \@xhline
}
\newcommand{\ie}{\emph{i.e.}\xspace}
\newcommand{\eg}{\emph{e.g.}\xspace}

\title{%
  \smash{\raisebox{-0.25em}{\includegraphics[height=1.5em]{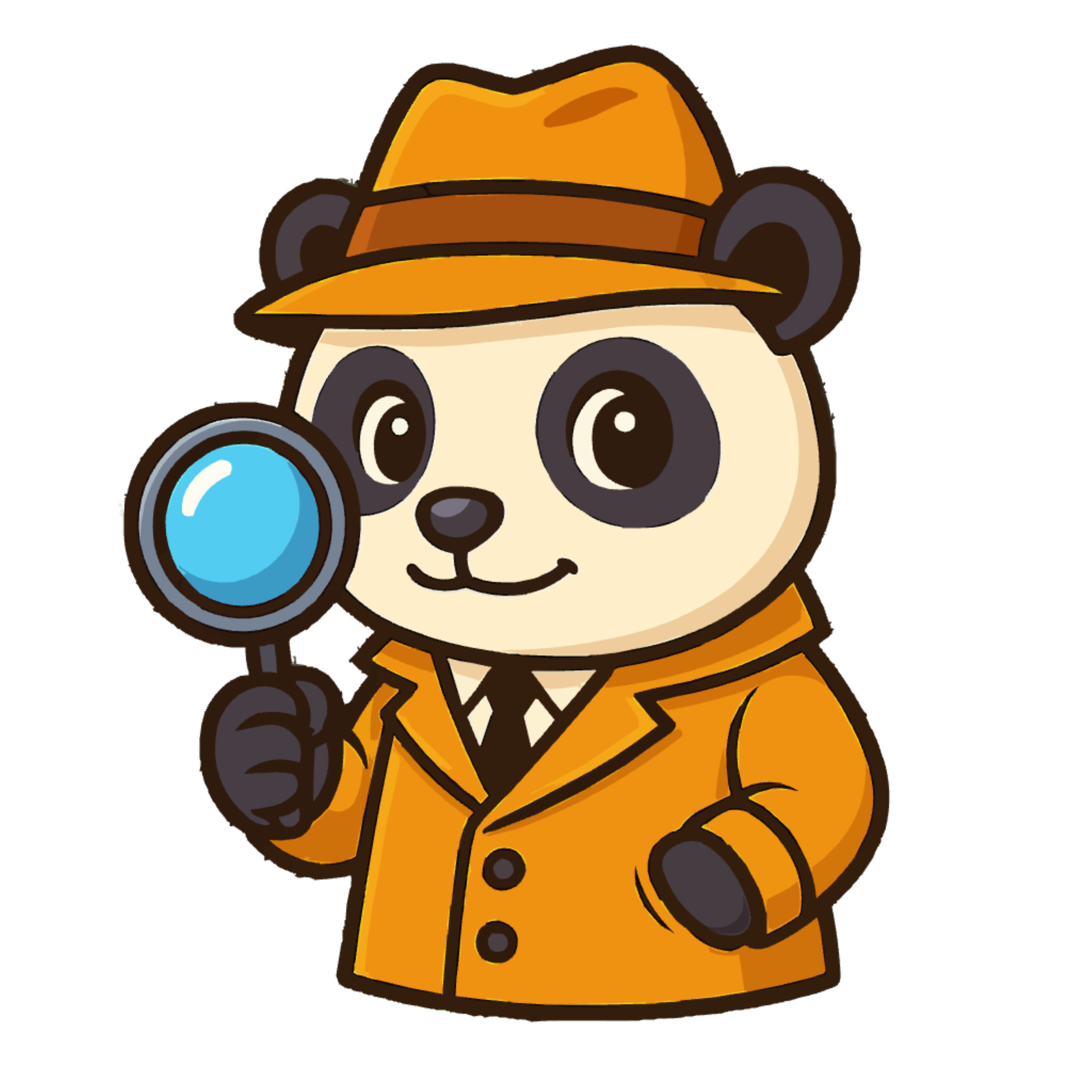}}}
  PANDA: Towards Generalist Video Anomaly Detection via Agentic AI Engineer
}

\makeatletter
\def\thanks#1{\protected@xdef\@thanks{\@thanks
        \protect\footnotetext{#1}}}
\makeatother

\author{%
  Zhiwei Yang\textsuperscript{1,2} \hspace{2em} Chen Gao\textsuperscript{2\dag} \hspace{2em} Mike Zheng Shou\textsuperscript{2\dag}\thanks{\dag Corresponding authors.} \\
 \textsuperscript{1}Xidian University \quad
\textsuperscript{2}Show Lab, National University of Singapore \\
}

\begin{document}

\maketitle

\begin{abstract}
Video anomaly detection (VAD) is a critical yet challenging task due to the complex and diverse nature of real-world scenarios. Previous methods typically rely on domain-specific training data and manual adjustments when applying to new scenarios and unseen anomaly types, suffering from high labor costs and limited generalization. Therefore, we aim to achieve generalist VAD, \ie, automatically handle any scene and any anomaly types without training data or human involvement. In this work, we propose PANDA, an agentic AI engineer based on MLLMs. Specifically, we achieve PANDA by comprehensively devising four key capabilities: (1) self-adaptive scene-aware strategy planning, (2) goal-driven heuristic reasoning, (3) tool-augmented self-reflection, and (4) self-improving chain-of-memory. Concretely, we develop a self-adaptive scene-aware RAG mechanism, enabling PANDA to retrieve anomaly-specific knowledge for anomaly detection strategy planning. Next, we introduce a latent anomaly-guided heuristic prompt strategy to enhance reasoning precision. Furthermore, PANDA employs a progressive reflection mechanism alongside a suite of context-aware tools to iteratively refine decision-making in complex scenarios. Finally, a chain-of-memory mechanism enables PANDA to leverage historical experiences for continual performance improvement. Extensive experiments demonstrate that PANDA achieves state-of-the-art performance in multi-scenario, open-set, and complex scenario settings without training and manual involvement, validating its generalizable and robust anomaly detection capability. Code is released at \textcolor{blue}{https://github.com/showlab/PANDA}.
\end{abstract}
\section{Introduction}
Video anomaly detection (VAD)~\cite{wu2024deep-8, hasan2016learning-12, benezeth2009abnormal-2, cong2011sparse-15} aims to identify abnormal or suspicious events in video streams, playing a vital role in a wide range of real-world applications such as intelligent surveillance~\cite{lu2013abnormal-18}, traffic monitoring~\cite{mahadevan2010anomaly-19}, autonomous driving~\cite{yao2022dota}, and industrial safety~\cite{hasan2016learning-12}. 

\begin{figure*}[t]
  \centering
   \includegraphics[width=1\linewidth]{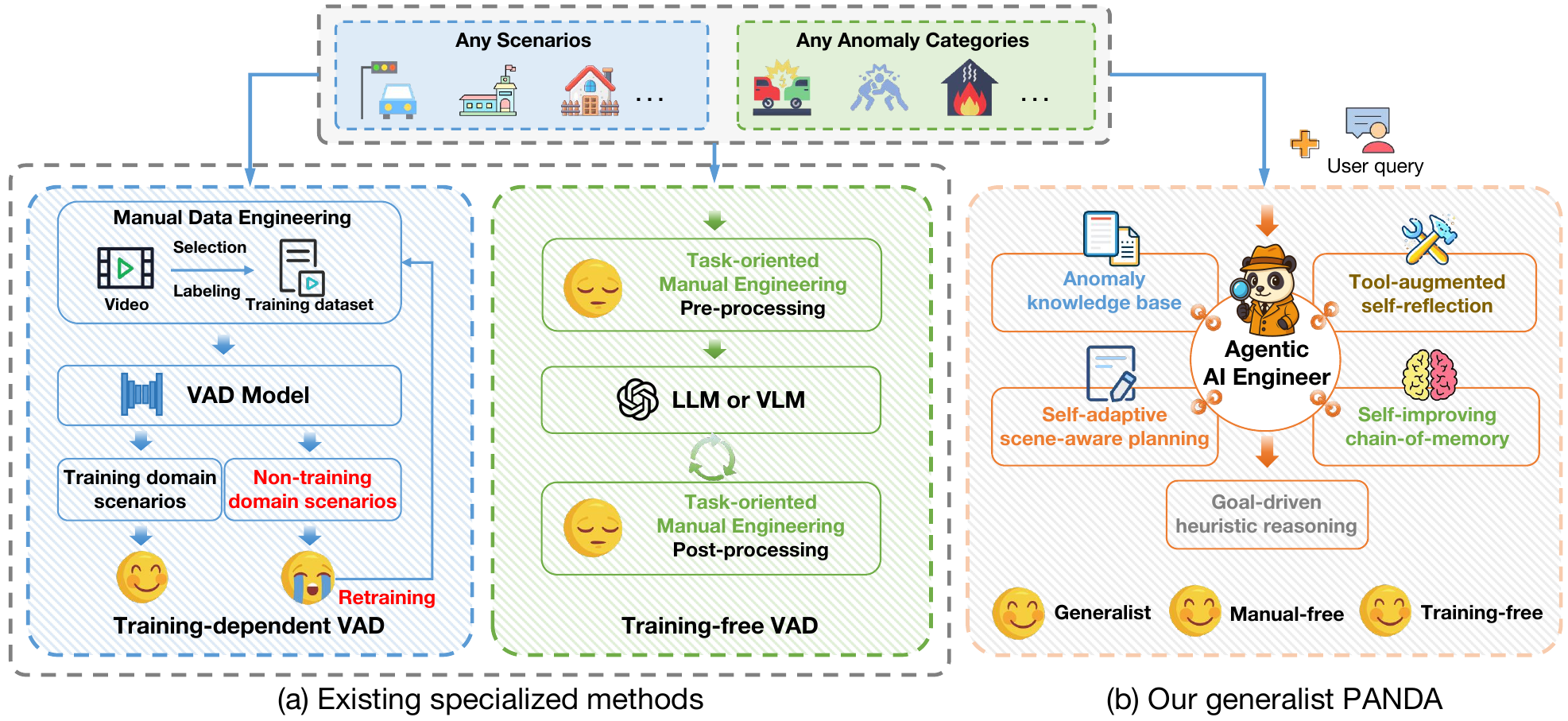}
    \caption{\textbf{PANDA vs. Existing specialized methods.} When facing arbitrary scenes and anomalies, PANDA can automatically adapt without the need for training refinements and manual adjustments, still achieving superior performance.}
    \vspace{-2mm}
   \label{fig:VAD_vs}
\end{figure*}

Existing VAD methods follow a specialist-oriented paradigm and require manual participation when deploying for new scenarios and anomalies. Broadly, they can be categorized into: training-dependent and training-free (Fig.~\ref{fig:VAD_vs}(a)). 
Specifically, training-dependent methods rely on newly annotated data to train models for each target scenario. The manual and training costs make such methods lack generalization and versatility.
Besides, training-free methods typically employ pre-trained large language models (LLMs) or vision-language models (VLMs) as the backbone, thereby eliminating the need for model training. However, they still depend heavily on manual engineering when deploying for new scenarios and anomalies, such as scenario-specific preprocessing steps, handcrafted prompt templates, rule curation, and post-processing. 
These static pipelines still lack adaptivity, making them brittle when confronted with uncertainty, long-term temporal dependencies, or complex, dynamic scenarios. Moreover, the hand-crafted nature restricts them from towarding generalist VAD.

To overcome the limitations of existing methods and free ourselves from the burden of domain-specific training and handcrafted pipeline design, our vision is to develop a general-purpose video anomaly detection method that can be self-adaptive to new scenarios and novel anomaly types without requiring any training data or manual pipeline assembly. The recent success of Multimodal Large Language Models (MLLMs) in a wide range of visual understanding tasks offers a promising foundation for realizing this vision. Therefore, we introduce {PANDA}, an agentic AI engineer for generalized VAD (Fig.~\ref{fig:VAD_vs}(b)). Drawing inspiration from how human engineers systematically analyze problems, adapt to complex environments,  and iteratively improve through tool use and experience, PANDA adaptively perceives the enviroment based on user-defined requirements, formulates detecion plans, perform goal-driven reasoning, invokes external tools to enhance decsion making, and continuously accumulates expeirence in memory for self-improvement. 

Technically, the proposed PANDA is distinguished by the following aspects. \textbf{\emph{(1) self-adaptive scene-aware strategy planning.}} Faced with a new scene or user-defined anomaly detection requirements, PANDA first conducts environment perception and understanding, then retrieves relevant anomaly rules from an anomaly knowledge database. Based on the environment context information, a scene-adaptive Retrieval Augmented Generation (RAG) mechanism is designed to construct tailored anomaly detection plans.
\textbf{\emph{(2) Goal-driven heuristic reasoning.}} PANDA injects task-specific prompts guided by latent anomaly cues, which steer the reasoning process toward more accurate and focused decision-making.
\textbf{\emph{(3) Tool-augmented self-reflection.}} PANDA iteratively assesses uncertainty and activates a suite of curated tools, such as object detection, image retrieve, or web search, to acquire additional information and resolve ambiguous decision-making.
\textbf{\emph{(4) Self-improving chain-of-memory.}} PANDA integrates historical experiences to justify current reasoning decisions or self-reflection. By progressively accumulating contextual cues across temporal spans, it enhances both the stability and accuracy of its decisions over time.
Taken together, PANDA embodies an agentic AI engineer that proactively perceives diverse environments, formulates adaptive strategies, performs goal-driven reasoning, and progressively improves through tool-augmented reflection and a chain-of-memory mechanism, enabling generalizable video anomaly detection across complex real-world scenarios.

Extensive experiments across multiple challenging benchmarks show that PANDA achieves {state-of-the-art performance} in multi-scenario, open-set, and complex scenario settings, \emph{without training and manual involvement}. 
These results highlight PANDA’s strong potential as an autonomous and general-purpose solution for real-world VAD.

\vspace{-2mm}
\section{Related Work}
\vspace{-2mm}
Video anomaly detection (VAD)~\cite{pang2020self-16, luo2017remembering-14, liu2018future-13, gong2019memorizing-11},  has long been a critical research topic in the computer vision field due to its wide range of real-world applications. Existing VAD methods are specialist-oriented and can be broadly categorized into training-dependent and training-free approaches.
\paragraph{Training-dependent VAD.}
These approaches rely on varying levels of annotated data and typically fall into three categories: semi-supervised VAD~\cite{hasan2016learning-12, lu2013abnormal-18, liu2018future-13, yang2023video-17, yang2022dynamic-15}, weakly-supervised VAD~\cite{yang2024text-21, wu2024vadclip-25, wu2020not-24, sultani2018real-22, wu2024weakly, lian2025vlial}, and instruction-tuned VAD~\cite{du2024uncovering-29, tang2024hawk-30, zhang2024holmes-07, zhang2024holmes-32, yang2025assistpda}. For example, Ristea et al.~\cite{ristea2024self} proposed an efficient anomaly detection model based on a lightweight masked autoencoder. Yang et al.~\cite{yang2024text-21} introduced a text prompt-driven pseudo-labeling and self-training framework for weakly-supervised VAD. Zhang et al.~\cite{zhang2024holmes-32} presented a model combining an anomaly-focused temporal sampler with an instruction-tuned MLLM to detect anomalies. While these training-dependent methods often perform well within the domain of the training data, they typically suffer from sharp performance degradation when deployed in out-of-distribution environments or faced with novel anomaly types. This limits their applicability in the open-world scenarios where anomalies are diverse, and context-sensitive.
\paragraph{Training-free VAD.}
Inspired by the recent success of LLMs~\cite{achiam2023gpt, team2023gemini} and VLMs~\cite{bai2025qwen2, team2024internvl2-05, Video-llava-03}, training-free VAD methods~\cite{zanella2024harnessing-28, yang2024follow-31} have gained increasing attention. These approaches aim to leverage the powerful prior knowledge embedded in foundation models without requiring domain-specific training. For instance, Zanella et al.~\cite{zanella2024harnessing-28} proposed the first language-model-based training-free VAD framework, which improves anomaly scoring by aligning cross-modal features between LLMs and VLMs while suppressing noisy captions. Yang et al.~\cite{yang2024follow-31} developed a rule-based anomaly inference framework by prompting LLMs to perform inductive and deductive reasoning over anomaly rules. Despite removing the need for training, these methods often rely on static prompting patterns and require substantial manual engineering (\eg, handcrafted pre/post-processing), which limits their adaptivity and robustness in complex, real-world scenarios.

Distinct from both paradigms above, PANDA is an agent-based framework that embodies the characteristics of an agentic AI engineer, which is capable of autonomously performing VAD without training and manual engineering when faced with various real-world scenarios. By incorporating a progressive reflection mechanism and a suite of perception-enhancing tools, PANDA can adaptively refine its predictions through self-reflection and tool invocation. This enables PANDA to dynamically handle diverse and challenging scenarios in the real world.
\section{Method}
\label{method}
\begin{figure*}[t]
  \centering
   \includegraphics[width=1\linewidth]{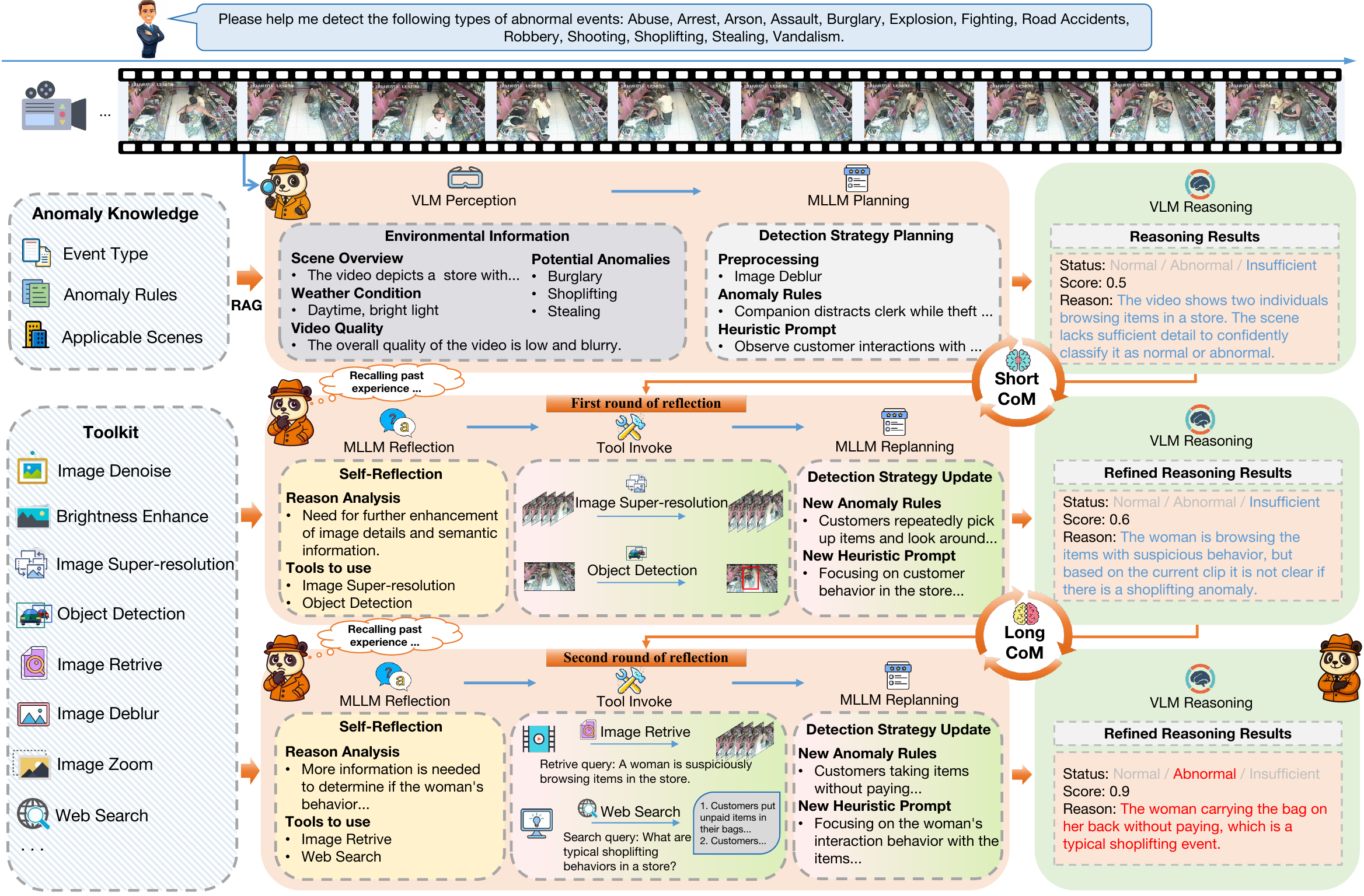}
   \caption{\textbf{Overview of the proposed PANDA.} As shown in the figure, upon receiving a user-defined query, PANDA first performs environment perception and plans a scene-adaptive detection strategy. PANDA then executes the plan with goal-driven heuristic reasoning. When encountering ambiguous cases, PANDA enters a reflection phase, revising its plan and invoking external tools to refine the decision. Throughout the process, PANDA maintains both short-term and long-term Chain-of-Memory (CoM), enabling it to accumulate experience and continually improve over time.}
   \label{fig:Illustration}
\end{figure*}

In this section, we present the core architecture and reasoning process of PANDA, an agentic AI engineer for generalized VAD. 
PANDA is designed to dynamically perceive diverse environments and perform progressive, tool-enhanced reasoning and self-refinement, as shown in Fig.~\ref{fig:Illustration}. 
It achieves this through four synergistic modules: (1) self-adaptive scene-aware strategy planning, (2) goal-driven heuristic reasoning, (3) tool-augmented self-reflection, and (4) self-improving chain-of-memory. 
\subsection{Self-adaptive Scene-aware Strategy Planning}
\label{planning}
To achieve VAD in general and unconstrained environments, it is essential to dynamically perceive the current video context and construct targeted detection strategies. 
Given the scene-dependence of many real-world anomalies and the variability in visual conditions, PANDA first performs self-adaptive perception of the input video to extract high-level environment contextual information.

\paragraph{Environmental Perception.}
Given a user-defined detection query \( \texttt{User}_{\text{query}} \) and an input video sequence \( V = \{f_1, f_2, \dots, f_N\} \) containing $N$ frames, PANDA uniformly samples \( M \) keyframes \( F = \{f_1, f_2, \dots, f_M\} \) and constructs a perception prompt $\texttt{Prompt}_{\text{perception}}$ combining \( F \) with the \( \texttt{User}_{\text{query}} \). This prompt is fed to a VLM, which returns structured environmental information:
\begin{equation}
\begin{aligned}
\texttt{EnvInfo} &= \text{VLM}(F, \texttt{Prompt}_{\text{perception}}) \\
         &= \{\texttt{Scene~Overview}, \texttt{Potential~Anomalies}, \\
         &\;\;\texttt{Weather~Condition}, \texttt{Video~Quality}\}
\end{aligned}
\end{equation}
Here, \texttt{Scene Overview} provides a high-level summary of the scene, including location type (\eg, street, shop, parking lot) and observed activities. \texttt{Potential Anomalies} refers to types of suspicious behaviors that may plausibly occur in the current scene context. \texttt{Weather Condition} captures attributes such as time of day (day/night) and weather (\eg, sunny, rainy). \texttt{Video Quality} summarizes resolution and clarity (\eg, low-resolution, blurred, noisy).

\paragraph{RAG-Based Strategy Planning.}
With the structured environment context in hand, PANDA proceeds to plan its detection strategy. To avoid hallucinations and improve reliability, this planning is performed via retrieval-augmented generation (RAG)~\cite{lewis2020retrieval}, driven by a multimodal large language model (MLLM). First, based on \( \texttt{User}_{\text{query}} \), PANDA constructs a knowledge base prompt \( \texttt{Prompt}_{\text{know}} \) and then generates a structured general anomaly knowledge base using the MLLM:
\begin{equation}
\kappa_a = \text{MLLM}(\texttt{User}_{\text{query}}, \texttt{Prompt}_{\text{know}}) = \{\texttt{Event~Type}, \texttt{Anomaly~Rules}, \texttt{Application~Scenes}\}.   
\end{equation}
Here, \texttt{Event Type} indicates anomaly categories specified by the user. \texttt{Anomaly Rules} are detection rules associated with each anomaly type. \texttt{Application Scenes} are contextual environments where anomalies are likely to occur.

For each anomaly type, we predefine \( H \) rule-scene pairs to form the knowledge base. PANDA then uses the perceived \texttt{EnvInfo} as a query to retrieve the top-\( k \) most relevant anomaly rules:
\begin{equation}
\texttt{Rules}_a = \texttt{RetrieveTopK}(\texttt{EnvInfo}, \kappa_a).
\end{equation}

Finally, PANDA integrates the \( \texttt{User}_{\text{query}} \), $\texttt{EnvInfo}$, and $\texttt{Rules}_a$ to construct a planning prompt \( \texttt{Prompt}_{\text{plan}} \), which is passed to the MLLM to generate the detection strategy plan:

\begin{equation}
\begin{aligned}
\texttt{Plan}_{\texttt{strategy}} &= \text{MLLM}(\texttt{Prompt}_{\text{plan}}) \\
                &= \{\texttt{Preprocessing}, \texttt{Potential~Anomalies}, \\
                &\;\;\texttt{Heuristic~Prompt}\}.
\end{aligned}
\end{equation}
Here, \texttt{Preprocessing} specifies optional visual enhancement steps (\eg, brightness adjustment, denoising, super-resolution). \texttt{Potential Anomalies} refines the anomaly list based on rule relevance and scene understanding. \texttt{Heuristic Prompt} includes step-by-step reasoning instructions for each potential anomaly, enabling the downstream inference module to perform structured, chain-of-thought analysis.

By integrating self-adaptive environment perception and RAG-enhanced strategy planning, PANDA ensures that subsequent anomaly reasoning is goal-driven and context-aware, significantly improving robustness in open-world settings.

\subsection{Goal-Driven Heuristic Reasoning}
\label{subsection:Reasoning}
The reasoning module serves as the core component of PANDA for analyzing video anomaly events. PANDA supports both offline and online inference modes. In this section, we focus on the offline setting, while the implementation details section will describes the online mode settings.

Under the guidance of the detection strategy plan constructed in subsection~\ref{planning}, PANDA performs goal-driven heuristic reasoning using a VLM. Given the \( \texttt{User}_{\text{query}} \), a clip-level video segment \( V_{\text{clip}} = \{c_1, c_2, \dots, c_T\} \) (each video clip $c_t$ contains $s$ video frames), and the $\texttt{Plan}_{\texttt{strategy}}$, PANDA first applies the preprocessing tools specified in $\texttt{Plan}_{\texttt{strategy}}$ to obtain an enhanced video clip:
\begin{equation}
\widetilde{V}_{\text{clip}}= \text{Preprocessing} (V_{\text{clip}}) = \{\widetilde{c}_1, \widetilde{c}_2, \dots, \widetilde{c}_T\}.
\end{equation}
Next, PANDA constructs a reasoning prompt based on the $\texttt{Plan}_{\texttt{strategy}}$: 

\texttt{Prompt\textsubscript{reasoning} = \{\ensuremath{\text{Memory}^{\text{$l$-steps}}_{\text{text}}}, Potential~Anomalies, $\texttt{Rules}_a$, Heuristic~Prompt, Enhancement~and~Reflection~Info\}}.

The fields \texttt{Potential Anomalies}, $\texttt{Rules}_a$, and \texttt{Heuristic Prompt} are directly inherited from the planning stage. The \texttt{Enhancement and Reflection Info} field incorporates information produced during the self-reflection stage (To be described in subsection~\ref{subsection:reflection}), including tool-based refinements and updated anomaly rules and heuristic prompts. To enhance temporal awareness, PANDA equips a short-term memory component \( \texttt{Memory}^{l\text{-steps}}_{\text{text}} \), which records the past \( l \) reasoning steps as textual memory. In addition to textual memory, PANDA also maintains a corresponding visual memory stream \( \texttt{Memory}^{l\text{-steps}}_{\text{visual}} \), which stores visual frames aligned with the latest \( l \) steps, allowing the model to access fine-grained visual cues during inference.

Finally, driven by the potential anomaly targets and enriched contextual knowledge, PANDA performs heuristic reasoning with the following formulation:

\begin{equation}
\begin{aligned}
\texttt{Result}_{\text{reasoning}} 
   &= \text{VLM}(\widetilde{c}_t, \texttt{Memory}^{l\text{-steps}}_{\text{visual}}, \texttt{Prompt}_{\text{reasoning}}) \\
   &= \{\texttt{Status}: \texttt{Normal} / \texttt{Abnormal} / \texttt{Insufficient}, \\
   &\;\;\texttt{Score} \in [0,1],\ \texttt{Reason}: \langle \cdot \rangle\}.
\end{aligned}
\end{equation}

Here, \texttt{Status} indicates the result of the VLM judgment: 
\texttt{Normal} indicates the clip is confidently classified as non-anomalous, 
\texttt{Abnormal} denotes strong evidence of anomaly, and 
\texttt{Insufficient} suggests the current information is inadequate to make a definitive judgment. \texttt{Score} is the probability of the existence of an abnormal event for the clip corresponding to each status. \texttt{Reason} is the reason for the status judgment given by the VLM.
When the result is \texttt{Insufficient}, PANDA will trigger the reflection mechanism to gather additional context or observations before re-entering the reasoning loop.

\subsection{Tool-Augmented Self-Reflection}
\label{subsection:reflection}
In complex scenarios, PANDA may not be able to make a clear decision on whether a video segment is normal or abnormal. In such ambiguous cases, it returns an \texttt{Insufficient} status, which triggers the reflection module. PANDA adopts a tool-augmented self-reflection mechanism enhanced by a specialized set of tools \(\tau = \{\texttt{tool}_1, \texttt{tool}_2, \dots, \texttt{tool}_n\}\) for visual content enhancement and auxiliary analysis, including image deblurring, denoising, brightness enhancement, image retrieval, object detection, and web search, etc. These tools assist in gathering additional evidence to support the decision-making process.

\paragraph{Experience-Driven Reflection.} Given an \texttt{Insufficient Reason} from the current reasoning step, PANDA first queries its long chain-of-memory (\texttt{Long CoM}, will be introduced in \ref{subsection:CoM}) to retrieve the most similar history reflection cases:
\begin{equation}
\texttt{Experience}_{\text{reflection}} = \texttt{RetrieveTop1}(\texttt{Insufficient Reason}, \texttt{Long CoM}).
\end{equation}

PANDA then constructs a reflection prompt using video context information, including the \(\texttt{User}_{\text{query}}\), \(\texttt{EnvInfo}\), $\texttt{Plan}_{\texttt{strategy}}$, \(\texttt{Rules}_a\), short chain-of-memory (short CoM), \(\texttt{Insufficient Reason}\), and \(\texttt{Experience}_{\text{reflection}}\):
$
\texttt{Prompt}_{\text{reflection}} = \{\texttt{User}_{\text{query}}, \texttt{EnvInfo}, \texttt{Plan}_{\texttt{strategy}}, \texttt{Rules}_a,  \text{short CoM}, \\
\texttt{Insufficient Reason}, \texttt{Experience}_{\text{reflection}}\}.
$
This prompt is fed into the  MLLM to analyze the cause of uncertainty and recommend an appropriate reflection plan:

\begin{equation}
\begin{aligned}
\texttt{Result}_{\text{reflection}} 
   &= \text{MLLM}(\texttt{Prompt}_{\text{reflection}}) \\
   &= \{\texttt{"Insufficient Reason"}: \langle\cdot\rangle, \\
   &\;\;\texttt{"Tools to Use"}: [\{\texttt{tool}_1, \texttt{params}\}, \dots, \{\texttt{tool}_n, \texttt{params}\}], \\
   &\;\;\texttt{"New Anomaly Rule"}: \langle\cdot\rangle, \\
   &\;\;\texttt{"New Heuristic Prompt"}: \langle\cdot\rangle \}.
\end{aligned}
\end{equation}

Here, \texttt{Insufficient Reason} refers to the underlying cause of decision uncertainty, inferred by the MLLM in conjunction with VLM output and contextual information such as environmental cues and anomaly rules. \texttt{Tools to Use} specifies the names of tools used for information enhancement and their corresponding parameters. \texttt{New Anomaly Rule} and \texttt{New Heuristic Prompt} represent the updated anomaly rule and the reformulated heuristic prompt, respectively. 


\paragraph{Tool Invocation.} PANDA executes the tool functions suggested in the reflection result to enhance both visual and semantic information. The tool invocation process is formulated as:

\begin{equation}
\begin{aligned}
\texttt{Result}_{\text{tool\_augmented}} 
   &= \texttt{ToolInvoke}(\texttt{tool}_1, \dots, \texttt{tool}_n) \\
   &= \{\texttt{Text Enhancement Info}, \\
   &\;\;\texttt{Visual Enhancement Info} = \widehat{c}_t \cup c_s \}.
\end{aligned}
\end{equation}

Here, \texttt{Text Enhancement Info} includes summaries from tool outputs (\eg, detected objects, web search results), while \texttt{Visual Enhancement Info} includes processed video clip \(\widehat{c}_t\) and retrieved historical keyframe set \(c_s=\{f_1, f_2, ..., f_s \}\).

\paragraph{Refined Reasoning.} PANDA updates the reasoning prompt with the newly acquired textual cues:
\begin{equation}
\small
\begin{aligned}
\texttt{Prompt}^{\text{refined}}_{\text{reasoning}} 
   &= \texttt{Prompt}_{\text{reasoning}} \cup \{ \\
   &\;\;\texttt{Text Enhancement Info},\ \texttt{New Anomaly Rule}, \\
   &\;\;\texttt{New Heuristic Prompt} \}
\end{aligned}
\end{equation}
and re-reasoning the enhanced video clip input:
\begin{equation}
\texttt{Result}^{\text{reflection}}_{\text{reasoning}} = \text{VLM}(\widehat{c}_t \cup c_s,\ \texttt{Memory}^{l\text{-steps}}_{\text{visual}},\ \texttt{Prompt}^{\text{refined}}_{\text{reasoning}}).
\end{equation}
If the returned status is \texttt{Normal} or \texttt{Abnormal}, PANDA resumes reasoning at the next timestep. If the status remains \texttt{Insufficient}, reflection is re-triggered. To prevent infinite loops, we limit the number of reflection rounds to \(r\). If after \(r\) rounds the result is still \texttt{Insufficient}, PANDA assigns a default anomaly score corresponding to the \texttt{Insufficient} status and skips the current segment and continues next timestep.

\begin{figure*}[t]
  \centering
   \includegraphics[width=1\linewidth]{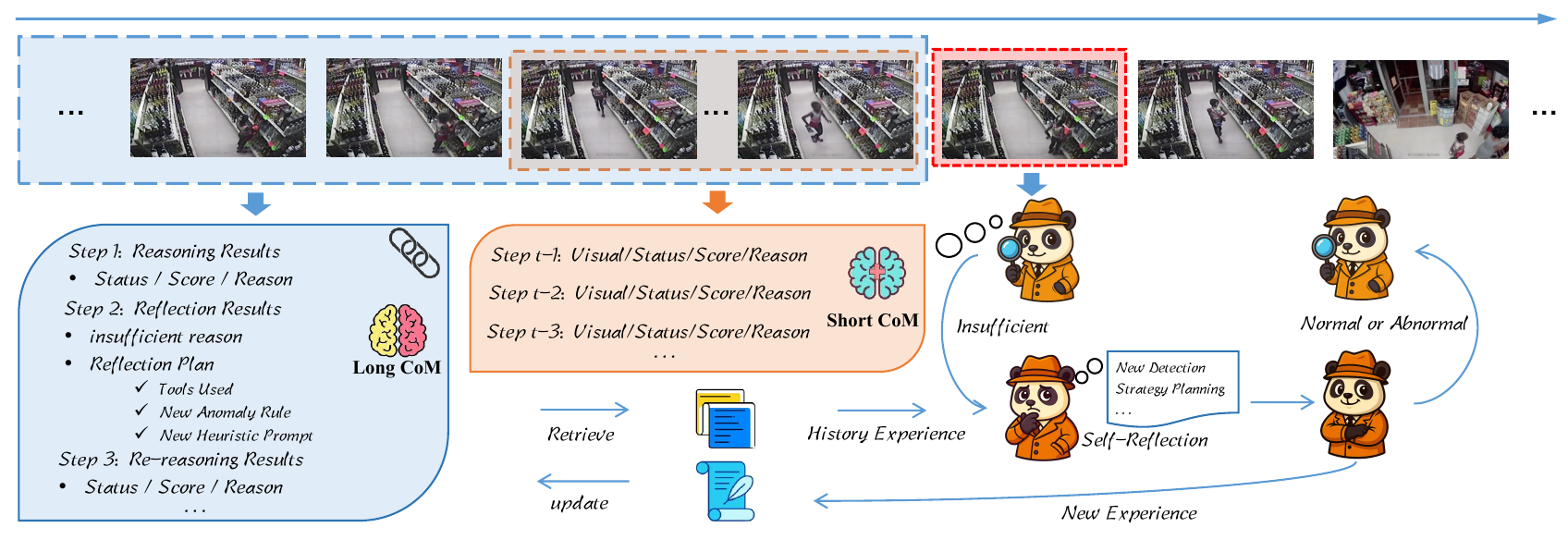}
   \caption{Illustration of Self-Improving Chain-of-Memory.}
   \label{fig:CoM-Illustration}
\end{figure*}

\subsection{Self-Improving Chain-of-Memory}
\label{subsection:CoM}
To enable PANDA to become increasingly "smarter" over time by accumulating experience through the iterative cycle of reasoning, reflection, and refined reasoning, PANDA equips a self-improving chain-of-memory (CoM) mechanism  as shown in Fig.~\ref{fig:CoM-Illustration}. This mechanism enhances both long-term context awareness and consistency in decision-making across video sequences.
The CoM comprises two components: short chain-of-memory (short CoM) and long chain-of-memory (long CoM). 

\paragraph{Short CoM.} In the reasoning stage, short CoM includes both the textual reasoning trace \(\texttt{Memory}_{\text{text}}^{l\text{-steps}}\) and its visual counterpart \(\texttt{Memory}_{\text{visual}}^{l\text{-steps}}\), as described in subsection~\ref{subsection:Reasoning}. In the reflection stage, short CoM is represented by the set of past reflection outputs:
$
\texttt{Result}^{\text{history}}_{\text{reflection}} = \{\texttt{Result}_{\text{reflection}}^{1}, \texttt{Result}_{\text{reflection}}^{2}, \dots,\texttt{Result}_{\text{reflection}}^{l}\}.
$

\paragraph{Long CoM.} PANDA also maintains a temporally evolving long CoM:
$
\texttt{Long-CoM} = \{M_1, M_2, \dots, M_T\},
$
where each memory unit \(M_t\) at time step \(t\) encapsulates three key outputs:
$
M_t = \{\texttt{Result}_{\text{reasoning}},\ \texttt{Result}_{\text{reflection}},\ \texttt{Result}_{\text{reasoning}}^{\text{refined}}\}.
$
This structure ensures that PANDA retains a complete trace of all decision stages—initial reasoning, reflective analysis, and post-reflection decisions. At the start of a video, LongCoM is empty by design, and PANDA relies on ShortCoM’s local window memory for initial reasoning and reflection. As more clips are processed, LongCoM gradually accumulates traces, supporting memory-consistent reasoning and reflection planning. With this self-improving chain-of-memory, PANDA leverages accumulated historical experience to inform both reasoning and reflection, leading to progressively more stable and accurate anomaly detection over time.
\section{Experiments}
\subsection{Experiment Setup}
\label{exp}
\paragraph{Datasets.}
We evaluate PANDA on four benchmarks: UCF-Crime~\cite{sultani2018real-22}, XD-Violence~\cite{wu2020not-24}, UBnormal~\cite{acsintoae2022ubnormal}, and CSAD, which represent three distinct settings—multi-scenario (UCF-Crime and XD-Violence), open-set (UBnormal), and complex scenario (CSAD). \textbf{UCF-Crime} is a large-scale dataset comprising 1,900 long, untrimmed real-world surveillance videos. It covers 13 types of abnormal events such as fighting, abuse, stealing, arson, robbery, and traffic accidents. The training set includes 800 normal and 810 abnormal videos, while the test set consists of 150 normal and 140 abnormal videos. \textbf{XD-Violence} is another large-scale dataset focused on violence detection. It contains 4,754 videos collected from surveillance video, movies, and CCTV sources, encompassing 6 categories of anomaly events. The training and test sets include 3,954 and 800 videos, respectively. \textbf{UBnormal} is a synthetic open-set video anomaly detection dataset with a total of 543 videos. It defines 7 categories of normal events and 22 types of anomalies. Notably, 12 anomaly categories in the test set are unseen during training, making it a challenging benchmark for evaluating generalization under open-set conditions. \textbf{CSAD} is a complex-scene anomaly detection benchmark constructed in this work. It consists of 100 videos (50 normal and 50 abnormal), sampled from UCF-Crime, XD-Violence, and UBnormal. CSAD includes videos with challenging conditions such as low resolution, poor illumination, high noise levels, and long-range temporal anomalies. It is designed to assess model robustness in complex and degraded environments.

\paragraph{Evaluation Metrics.}
Following the previous methods~\cite{wu2020not-24, sultani2018real-22}, we report the Area Under the Curve (AUC) of the frame-level receiver operating characteristic for UCF-Crime, UBnormal, and CSAD. For XD-Violence, we follow the evaluation criterion of average precision (AP) suggested by the work~\cite{wu2020not-24} to measure the effectiveness of our method.

\paragraph{Implementation Details.}
We adopt Langgraph~\cite{langchain2025langgraph} to build the whole agent framework and all experiments are implemented using PyTorch~\cite{imambi2021pytorch} on the A6000 GPU. We use Qwen2.5VL-7B~\cite{bai2025qwen2} as the VLM for perception and reasoning stages, and Gemini 2.0 Flash~\cite{team2023gemini} as the MLLM for planning and reflection. During the RAG process, the anomaly knowledge base and environment information are encoded via the all-MiniLM-L6-v2 model~\cite{reimers-2020-multilingual-sentence-bert}, with the knowledge base indexed using FAISS for efficient similarity retrieval. To improve the inference efficiency, the input video is sampled at 1 FPS, and a video clip of $s = 5$ frames is inferred at each time step. PANDA supports both offline and online reasoning modes. In offline reasoning mode, the perception phase is sampling $M=300$ frames uniformly for the whole video, while only the initial $M=10$ frames are sampled in online mode. The number of knowledge entries for each type of anomalous event in the anomaly knowledge base is $H=20$. The maximum number of reflection rounds \(r\) is set to 3. The short CoM length $l=5$ during the reasoning stage. We retrieve the top \(k = 5\) anomaly rules from the anomaly knowledge base for each user query. More implementation details, including prompt templates and tool usage, are provided in the supplementary material.

\begin{table}
\centering
\caption{Comparisons with previous state-of-the-art methods on different datasets. "Expl." stands for "Explanation", indicating whether the output results include interpretations of the detected anomalies. Methods categorized as "Semi", "Weak", or "Instru-Tuned" require training data to adapt to specific scenarios or anomaly types. 
}
\resizebox{\textwidth}{!}{
\renewcommand{\arraystretch}{1.35}
\setlength{\tabcolsep}{5pt}
\begin{tabular}{rl||c||c||c||c||cc||c||c}
\hline \thickhline

\rowcolor[HTML]{f8f9fa}
\multicolumn{2}{c||}{} & 
\multicolumn{1}{c||}{} &
\multicolumn{1}{c||}{} &
\multicolumn{1}{c||}{} &
\multicolumn{1}{c||}{} &
\multicolumn{2}{c||}{Multi-Scenario} & 
\multicolumn{1}{c||}{Open-Set} & 
\multicolumn{1}{c}{Complex Scenario} \\

\rowcolor[HTML]{f8f9fa}
\multicolumn{2}{c||}{\multirow{-2}{*}{\textbf{Methods}}} &
\multirow{-2}{*}{Supervision} & 
\multirow{-2}{*}{Expl.} & 
\multirow{-2}{*}{Manual-free} & 
\multirow{-2}{*}{Mode} &  
UCF (AUC\%) & XD (AP\%) & UB (AUC\%) & CSAD (AUC\%) \\
\hline
\hline

\rowcolor[HTML]{f8f9fa} 
\multicolumn{10}{l}{\emph{Specialized methods}} \\

AED-MAE\cite{ristea2024self} & $\!\!$\pub{CVPR2024} & Semi & {\ding{55}} & {\ding{55}} & Offline & - & - & 58.50 & - \\
STPAG\cite{rai2024video} & $\!\!$\pub{CVPR2024} & Semi & {\ding{55}} & {\ding{55}} & Offline & - & - & 57.98 & - \\
HL-Net\cite{wu2020not-24} & $\!\!$\pub{ECCV2020} & Weak & {\ding{55}} & {\ding{55}} & Offline & 82.44 & 73.67 & - & - \\
RTFM\cite{tian2021weakly-23} & $\!\!$\pub{ICCV2021} & Weak & {\ding{55}} & {\ding{55}} & Offline & 84.30 & 77.81 & 64.94 & - \\
UR-DMU\cite{zhou2023dual} & $\!\!$\pub{AAAI2023} & Weak & {\ding{55}} & {\ding{55}} & Offline & 86.97 & 81.66 & 59.91 & - \\
VadCLIP\cite{wu2024vadclip-25} & $\!\!$\pub{AAAI2024} & Weak& {\ding{55}} & {\ding{55}} & Offline & 88.02 & 84.51 & - & - \\
TPWNG\cite{yang2024text-21} & $\!\!$\pub{CVPR2024} & Weak& {\ding{55}} & {\ding{55}} & Offline & 87.79 & 83.68 & - & - \\
VERA\cite{Ye_2025_CVPR} & $\!\!$\pub{CVPR2025} & Weak& {\ding{51}} & {\ding{55}} & Offline & 86.55 & 70.11 & - & 64.52 \\
Holmes-VAU\cite{zhang2024holmes-32} & $\!\!$\pub{CVPR2025} & Instru-Tuned & \ding{51} & {\ding{55}} & Offline & 88.96 & 87.68 & 56.77 & 72.47 \\
\cline{3-10}


ZS CLIP\cite{bai2025qwen2} & $\!\!$\pub{ICML2021} & Training-free & \ding{51} & {\ding{55}} & Offline & 53.16 & 17.83 & 46.2 & 32.45 \\
LLAVA-1.5\cite{bai2025qwen2} & $\!\!$\pub{CVPR2024} & Training-free & \ding{51} & {\ding{55}} & Offline & 72.84 & 50.26 & 53.71 & 47.78 \\
LAVAD \cite{zanella2024harnessing-28} & $\!\!$\pub{CVPR2024} & Training-free & \ding{51} & {\ding{55}} & Offline & 80.28 & 62.01 & 64.23 & 57.26 \\
AnomalyRuler \cite{yang2024follow-31} & $\!\!$\pub{ECCV2024} & Training-free & \ding{51} & {\ding{55}} & Offline & - & - & 71.90 & - \\
\hline

\rowcolor[HTML]{f8f9fa} 
\multicolumn{10}{l}{\emph{Generalized method}} \\

\multirow{2}{*}{\textbf{PANDA}} & 
\multirow{2}{*}{\textbf{(ours)}} & 
\multirow{2}{*}{Training-free} & 
\multirow{2}{*}{\ding{51}} & 
\multirow{2}{*}{\ding{51}} & Offline & \textbf{84.89} & \textbf{70.16} & \textbf{75.78} & \textbf{73.12} \\
 &  &  &  &  & Online & 82.57 & 63.57 & 72.41 & 71.25 \\
\hline
\end{tabular}

}
\label{tab:sota}
\end{table}

\subsection{Comparison with State-of-the-Art Methods}
Table~\ref{tab:sota} compares the performance of PANDA against state-of-the-art specialized VAD methods, including both training-dependent and training-free methods. As shown, PANDA significantly outperforms all existing training-free baselines across all four datasets, even under online settings.

On the UBnormal dataset, which adopts an open-set evaluation protocol where test-time anomalies are unseen during training, PANDA surpasses both training-dependent and training-free approaches. This highlights PANDA’s strong generalization capabilities.

PANDA also exhibits notable advantages on CSAD, the complex-scene benchmark introduced in this work, where traditional methods tend to fail under low-quality or temporally extended anomalies. PANDA’s superior performance across diverse datasets and conditions demonstrates its robustness and effectiveness as a general-purpose solution for real-world video anomaly detection.

\subsection{Analytic Results}
\paragraph{Analysis of reflection round $r$.}
Table~\ref{tab:ablation-multi}(a) shows the effect of varying the number of reflection rounds \(r\) on PANDA's performance. We observe that performance improves gradually when increasing \(r\) from 1 to 5. Although \(r=5\) yields a slight additional improvement compared to \(r=3\), it introduces more computational overhead due to repeated tool invocation and reasoning steps. To balance efficiency and effectiveness, we adopt \(r=3\) as the default setting in all experiments.

\paragraph{Analysis of rules number $k$.}
Table~\ref{tab:ablation-multi}(b) analyses the influence of the number of retrieved rules \(k\) used during RAG-based anomaly strategy planning. When too few rules are retrieved (\eg, \(k=1\)), the system lacks diverse contextual cues to support robust reasoning, resulting in performance degradation. Conversely, setting \(k\) too high may introduce noisy or irrelevant rules that dilute reasoning quality. Finally, PANDA achieves optimal performance when setting \(k=5\).

\paragraph{Analysis of short CoM length $l$.}
Table~\ref{tab:ablation-multi}(c) analyzes the impact of varying the short CoM length \(l\) during the reasoning phase. The best performance is achieved when \(l = 5\). When the memory length is reduced to \(l = 1\), performance drops noticeably due to insufficient temporal information, which limits the model’s ability to leverage recent reasoning traces. On the other hand, increasing the memory length to \(l = 9\) also leads to performance degradation, likely because excessive memory introduces historical noise that distracts from the current decision-making process.

\begin{table}
\centering
\caption{Key hyperparameter analyses on the UCF-Crime dataset.}
\resizebox{\textwidth}{!}{
\subfloat[Analysis of reflection round $r$. \label{tab:ablation:reward}]{
   \renewcommand{\arraystretch}{1.1}
    \setlength{\tabcolsep}{5pt}
   \begin{tabular}{c||c}
\hline \thickhline

 \rowcolor[HTML]{f8f9fa} \multicolumn{1}{c||}{Reflection} & \multicolumn{1}{c}{UCF-Crime} \\
\rowcolor[HTML]{f8f9fa} \multicolumn{1}{c||}{Round $r$} & \multicolumn{1}{c}{(AUC\%)}  \\\hline
\hline
1 & 83.83  \\
3 & 84.89  \\
5 & \textbf{84.91}  \\
\hline
\end{tabular}\hspace{9mm}
}
\subfloat[Analysis of rules number $k$. \label{tab:ablation:grpo}]
{
   \renewcommand{\arraystretch}{1.1}
    \setlength{\tabcolsep}{5pt}
   \begin{tabular}{c||c}
\hline \thickhline

 \rowcolor[HTML]{f8f9fa} \multicolumn{1}{c||}{Rules} & \multicolumn{1}{c}{UCF-Crime} \\
\rowcolor[HTML]{f8f9fa} \multicolumn{1}{c||}{Number $k$} & \multicolumn{1}{c}{(AUC\%)}  \\\hline
\hline
1 & 82.79  \\
5 & \textbf{84.89}  \\
9 & 83.92  \\
\hline
\end{tabular}\hspace{9mm}
}

\subfloat[Analysis of short CoM length $l$. \label{tab:ablation:grpo}]
{
   \renewcommand{\arraystretch}{1.1}
    \setlength{\tabcolsep}{5pt}
   \begin{tabular}{c||c}
\hline \thickhline

 \rowcolor[HTML]{f8f9fa} \multicolumn{1}{c||}{Short CoM} & \multicolumn{1}{c}{UCF-Crime} \\
\rowcolor[HTML]{f8f9fa} \multicolumn{1}{c||}{length $l$} & \multicolumn{1}{c}{(AUC\%)}  \\\hline
\hline
1 & 82.92  \\
5 & \textbf{84.89}  \\
9 & 84.03  \\
\hline
\end{tabular}\hspace{9mm}
}
}
\label{tab:ablation-multi}
\end{table}

\begin{table}
  \caption{
    {Ablation study on the capability of PANDA.}
    \textbf{Planning} refers to the self-adaptive scene-aware detection strategy planning, which contains detection planning, adaptive environment perception, and RAG with anomaly knowledge.
    \textbf{Reflection} denotes the tool-augmented self-reflection mechanism.
    \textbf{Memory} corresponds to the chain-of-memory module, encompassing both short-term and long-term components.
  }
  \label{tab:ablation_capability}
  \centering
  \footnotesize
  \renewcommand{\arraystretch}{1.35}
  \setlength{\tabcolsep}{5pt}
  \begin{tabular}{c||c||c||c}
    \hline \hline
    \rowcolor[HTML]{f8f9fa}
    \multicolumn{3}{c||}{\textbf{Key Capabilities of PANDA}} & \textbf{UCF-Crime (AUC\%)} \\
    \hline
    \rowcolor[HTML]{f8f9fa}
    \textbf{Planning} & \textbf{Reflection} & \textbf{Memory} & \textbf{Performance} \\
    \hline\hline
    \textcolor{red}{\ding{55}} & \textcolor{red}{\ding{55}} & \textcolor{red}{\ding{55}} & 75.25 \\
    \textcolor{green}{\ding{51}} & \textcolor{red}{\ding{55}} & \textcolor{red}{\ding{55}} & 80.37 (+5.12\%) \\
    \textcolor{green}{\ding{51}} & \textcolor{green}{\ding{51}} & \textcolor{red}{\ding{55}} & 82.63 (+2.26\%) \\
    \textcolor{green}{\ding{51}} & \textcolor{green}{\ding{51}} & \textcolor{green}{\ding{51}} & \textbf{84.89 (+2.26\%)} \\
    \hline
  \end{tabular}
\end{table}

\begin{figure*}[t]
  \centering
   \includegraphics[width=1\linewidth]{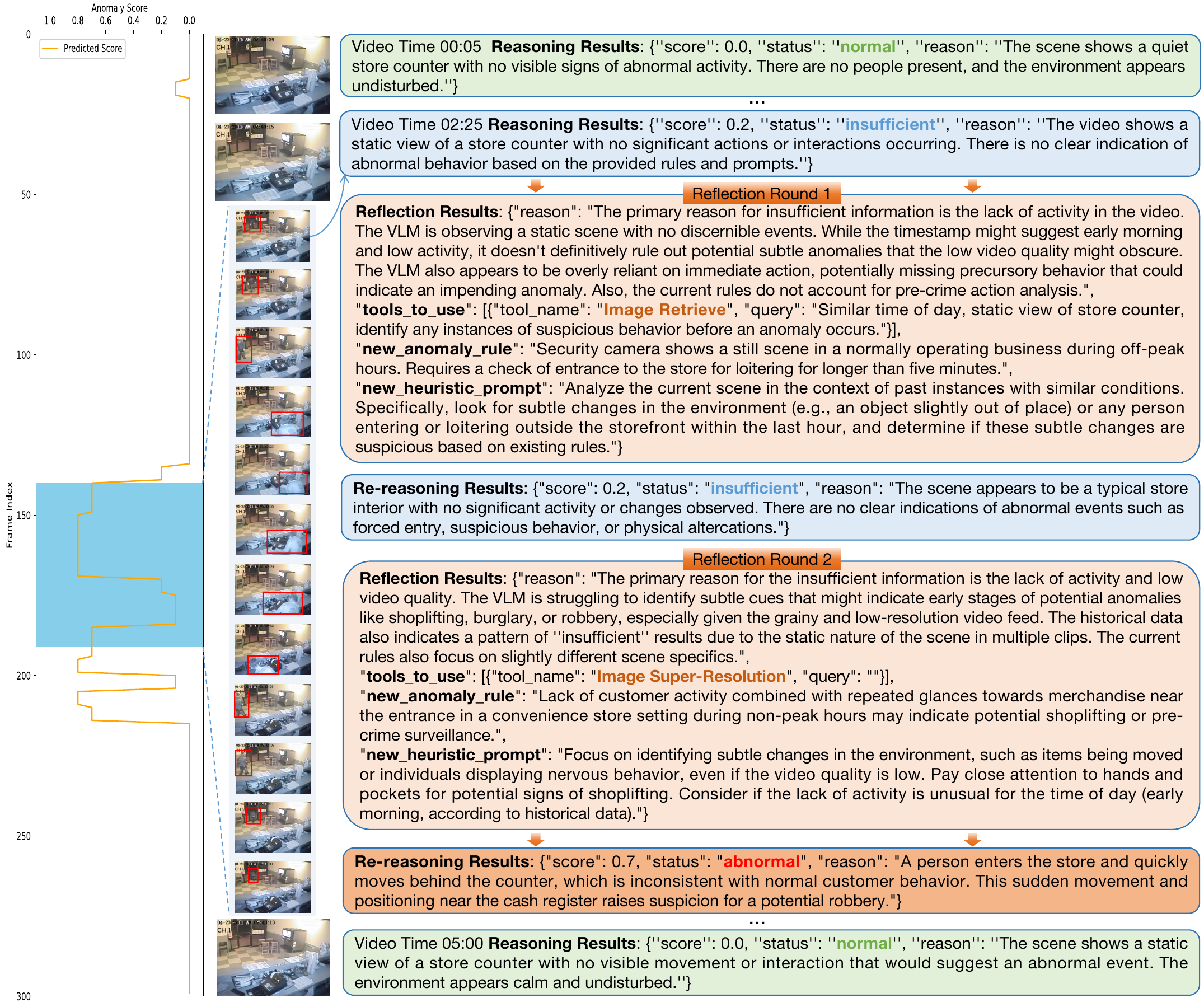}
   \caption{\textbf{Visualization of qualitative results on UCF-Crime.}
On the left is a visualization of the anomaly score curve. The right side shows the specific reasoning and reflection process of PANDA.}
   \label{fig:qualitative}
\end{figure*}

\paragraph{Ablation Study.}
Table~\ref{tab:ablation_capability} ~presents an ablation study examining the contribution of each core capability in PANDA, including self-adaptive scene-aware strategy planning (Planning), tool-augmented self-reflection (Reflection), and chain-of-memory (Memory). The third row serves as the baseline, where PANDA performs direct reasoning solely based on the user-defined query, without planning, reflection, and memory modules. As shown, the performance is relatively poor, with an AUC of 75.25\%. Equipping PANDA with the planning capability yields a substantial improvement of +5.12\% in AUC. This demonstrates the effectiveness of scene perception, rule retrieval via RAG, and context-aware strategy plans in inspiring the potential of PANDA. Adding the reflection module further improves performance by +2.26\%, suggesting that the self-reflection mechanism, enhanced by the integration of external tools, expands PANDA’s capability to resolve challenging and ambiguous cases. Finally, incorporating the memory mechanism results in another +2.26\% gain, validating the effectiveness of the chain-of-memory design. This module enables PANDA to accumulate experience across time and use it to refine decisions. In summary, each of PANDA’s capabilities plays a vital role in enabling generalizable and reliable anomaly detection. The synergistic integration of all modules empowers PANDA as a highly capable agentic AI engineer for generalized VAD. For the ablation study on the key components corresponding to each capability, please refer to the Subsection~\ref{Ablation of Key Component} in the supplementary material.

\subsection{Qualitative Results}
Figure~\ref{fig:qualitative} shows a visualized example from the UCF-Crime test set to illustrate PANDA’s reasoning-reflection process. The left side of the figure shows the anomaly score curve over time. On the right, we visualize PANDA’s internal reasoning and reflection process. When the model encounters uncertainty and cannot confidently determine whether an anomaly is present, it transitions into the reflection phase. PANDA first analyzes the reason behind the insufficient status from the reasoning stage, and then invokes external tools to acquire complementary information to support decision-making.
This case highlights PANDA’s capacity for progressive self-refinement and dynamic tool invocation, demonstrating its effectiveness in tackling complex, real-world video anomaly detection scenarios. For more visualization samples, please refer to the Supplementary Material.

\section{Conclusion}
In this work, we presented PANDA, an agentic AI engineer for generalized VAD that eliminates the need for training data or manually crafted pipelines when faced with various real-world scenarios. PANDA integrates four core capabilities: self-adaptive scene-aware strategy planning, goal-driven heuristic reasoning, tool-augmented self-reflection, and the self-improving chain-of-memory. These capabilities work in concert to enable PANDA to adaptively detect anomalies across diverse, dynamic, and previously unseen environments. Our extensive experiments across multiple benchmarks, including multi-scene, open-set, and complex scenarios, validate PANDA’s strong generalization ability and robust performance without any training. These findings highlight PANDA’s potential as a generalist VAD solution for real-world scenes.

\section*{Acknowledgement}
This research is supported by the National Research Foundation, Singapore under its AI Singapore Programme (AISG Award No: AISG3-RP-2022-030).

\clearpage
{
\small
\bibliographystyle{unsrt}
\bibliography{egbib}






\clearpage
\appendix

\section*{Technical Appendices and Supplementary Material}
\section{Overview of Technical Appendices and Supplementary Material}
\begin{figure*}[t]
  \centering
   \includegraphics[width=1\linewidth]{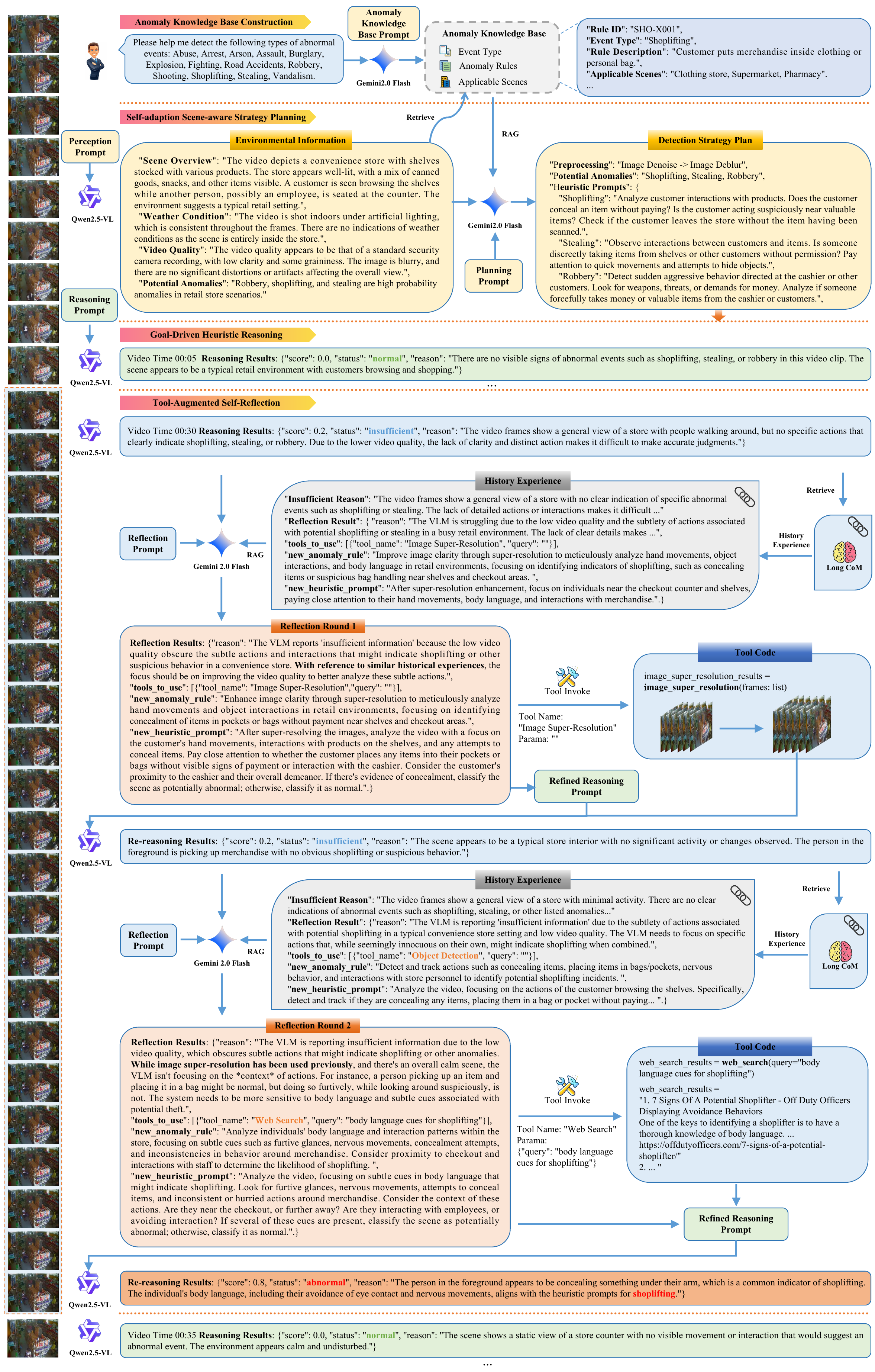}
    \caption{\textbf{Visualization of the PANDA detailed pipeline.} We show a more detailed pipeline for PANDA by visualizing the results of a test sample on UCF-Crime.}
   \vspace{-3mm}
   \label{fig:Vis_detail_method}
\end{figure*}

This technical appendices and supplementary material provide additional information not included in the main paper. Specifically:

\begin{itemize}
    \item \textbf{Section~\ref{Additional Methodological and Experimental Details}} offers further clarification on methodological and experimental details.
    \item \textbf{Section~\ref{Additional Experiments}} presents additional ablation studies and parameter analysis experiments.
    \item \textbf{Section~\ref{Toolset}} details the toolset employed by PANDA.
    \item \textbf{Section~\ref{System Prompt}} shows the prompts used at each stage of the PANDA framework.
    \item \textbf{Section~\ref{Additional visualization results}} provides additional visualizations of qualitative results across various datasets.
    \item \textbf{Section~\ref{Discussions}} discusses the current limitations, broader impacts, and future work directions.
\end{itemize}

\section{Additional Methodological and Experimental Details}
\label{Additional Methodological and Experimental Details}

\subsection{Visualization of the PANDA Detailed Pipeline}
Figure \ref{fig:Vis_detail_method} presents a visualized case study on a test example from the UCF-Crime dataset, illustrating the detailed execution process of PANDA across its core components: anomaly knowledge base construction, self-adaptive scene-aware strategy planning, goal-driven heuristic reasoning, tool-augmented self-reflection, and the self-improving chain-of-memory mechanism.

\subsection{Clarification on Evaluation Modes and SOTA Comparison}

In Table~1 of the main paper, we distinguish between \textit{offline} and \textit{online} settings based on whether future information is utilized when reasoning over a given frame or clip. If future information is accessed, the method is considered offline; otherwise, it falls under the online setting. For offline evaluation, we follow the SOTA methods AED-MAE, LAVAD, and AnomalyRuler, which are compared in Table~1 of the main paper, and apply temporal smoothing (mean filtering, window size=10) on the final anomaly scores.

Among the compared methods, Holmes-VAU is a fine-tuned VLM-based approach that leverages detailed anomaly annotations via instruction tuning. It was originally evaluated only on UCF-Crime and XD-Violence. Its results on UBnormal and CSAD are reproduced by us, without any re-training, using its publicly released model. As seen in Table~1 of the main paper, PANDA significantly outperforms Holmes-VAU on UBnormal, an open-set dataset that includes unseen anomaly types. While Holmes-VAU slightly surpasses PANDA in online mode on CSAD, this is primarily due to CSAD including a large number of videos derived from UCF-Crime and XD-Violence—the original training sets of Holmes-VAU. Notably, PANDA still achieves superior performance under the offline setting, underscoring its strong generalization capability. These results demonstrate the limitations of relying solely on fine-tuned VLMs when facing domain shift and complex real-world conditions. Additionally, we report new evaluation results for three prominent training-free baselines: ZS-CLIP, LLaVA-1.5, and LAVAD, on UBnormal and CSAD. PANDA consistently and substantially outperforms all of them across both datasets. Together, these experimental results reinforce the strength of PANDA as a generalist, fully automated VAD agent, capable of adapting its reasoning to scene-specific conditions without supervision or hand-crafted engineering.

\section{Additional Experiments}
\label{Additional Experiments}

\subsection{Ablation of Key Component}
\label{Ablation of Key Component}
Table~\ref{tab:ablation} presents the results of an ablation study evaluating the contribution of PANDA's six core components. The third row in the table corresponds to a baseline that directly queries the VLM using only the user-defined anomaly description without leveraging any PANDA modules. This baseline yields notably poor performance, confirming that naive prompting alone is insufficient. As more modules are incrementally added—namely, detection strategy planning, adaptive scene perception, RAG with anomaly knowledge, self-reflection, short CoM, long CoM—the performance steadily improves. When all six components are combined, PANDA achieves its best overall performance. These results demonstrate that each individual module contributes positively to the final performance and validates the effectiveness of our whole PANDA framework design.

\subsection{Impact of Different MLLMs}

Table~\ref{tab:Impact of different MLLMs} compares the performance of PANDA when integrated with different multi-modal large language models (MLLMs). GPT-4o and Gemini 2 Flash represent proprietary models, while DeepSeek-V3 and Qwen2.5-72B are open-source alternatives. As shown, GPT-4o achieves the highest performance. However, we adopt Gemini 2 Flash in our main pipeline due to its strong trade-off between performance and cost-effectiveness.

Notably, although Qwen2.5-72B yields the lowest performance among the compared models, it remains significantly superior to prior training-free baselines. Given its open-source nature and ease of local deployment, it serves as a practical and scalable option for resource-constrained scenarios.

\begin{table}
\centering
\caption{Ablation results of PANDA component.}
\resizebox{\textwidth}{!}{
\renewcommand{\arraystretch}{1.35}
\setlength{\tabcolsep}{5pt}
\begin{tabular}{c||c||c||c||c||c||c}
\hline \thickhline

\rowcolor[HTML]{f8f9fa}
\multicolumn{6}{c||}{\textbf{Key Components of PANDA}} & \textbf{Dataset} \\
\hline

\rowcolor[HTML]{f8f9fa}
\textbf{Detection Strategy Planning} & 
\textbf{Self-Adaption Scene-Aware} & 
\textbf{RAG with Anomaly Knowledge} & 
\textbf{Self-Reflection} & 
\textbf{Short CoM} & 
\textbf{Long CoM} & 
\textbf{UCF-Crime(AUC\%)} \\
\hline\hline 

\textcolor{red}{\ding{55}} & \textcolor{red}{\ding{55}} & \textcolor{red}{\ding{55}} & \textcolor{red}{\ding{55}} & \textcolor{red}{\ding{55}} & \textcolor{red}{\ding{55}} & 75.25 \\
\textcolor{green}{\ding{51}} & \textcolor{red}{\ding{55}} & \textcolor{red}{\ding{55}} & \textcolor{red}{\ding{55}} & \textcolor{red}{\ding{55}} & \textcolor{red}{\ding{55}} & 77.01 \\
\textcolor{green}{\ding{51}} & \textcolor{green}{\ding{51}} & \textcolor{red}{\ding{55}} & \textcolor{red}{\ding{55}} & \textcolor{red}{\ding{55}} & \textcolor{red}{\ding{55}} & 78.92 \\
\textcolor{green}{\ding{51}} & \textcolor{green}{\ding{51}} & \textcolor{green}{\ding{51}} & \textcolor{red}{\ding{55}} & \textcolor{red}{\ding{55}} & \textcolor{red}{\ding{55}} & 80.37 \\
\textcolor{green}{\ding{51}} & \textcolor{green}{\ding{51}} & \textcolor{green}{\ding{51}} & \textcolor{green}{\ding{51}} & \textcolor{red}{\ding{55}} & \textcolor{red}{\ding{55}} & 82.63 \\
\textcolor{green}{\ding{51}} & \textcolor{green}{\ding{51}} & \textcolor{green}{\ding{51}} & \textcolor{green}{\ding{51}} & \textcolor{green}{\ding{51}} & \textcolor{red}{\ding{55}} & 83.94 \\
\textcolor{green}{\ding{51}} & \textcolor{green}{\ding{51}} & \textcolor{green}{\ding{51}} & \textcolor{green}{\ding{51}} & \textcolor{green}{\ding{51}} & \textcolor{green}{\ding{51}} & \textbf{84.89} \\
\hline
\end{tabular}
}
\label{tab:ablation}
\end{table}

\begin{table}
\centering
\caption{Additional Experiments.}
\resizebox{\textwidth}{!}{
\subfloat[Impact of different MLLMs. \label{tab:Impact of different MLLMs}]{
   \renewcommand{\arraystretch}{1.1}
    \setlength{\tabcolsep}{5pt}
   \begin{tabular}{c||c}
\hline \thickhline

 \rowcolor[HTML]{f8f9fa} \multicolumn{1}{c||}{Different} & \multicolumn{1}{c}{UCF-Crime} \\
\rowcolor[HTML]{f8f9fa} \multicolumn{1}{c||}{MLLMs} & \multicolumn{1}{c}{(AUC\%)}  \\\hline
\hline
Qwen2.5-72B & 84.03  \\
DeepseekV3 & 84.72  \\
GPT4o & \textbf{84.97}  \\
Gemini 2 Flash & 84.89  \\
\hline
\end{tabular}\hspace{9mm}
}
\subfloat[Effect of input clip length $s$. \label{tab:Effect of input clip length}]
{
   \renewcommand{\arraystretch}{1.1}
    \setlength{\tabcolsep}{5pt}
   \begin{tabular}{c||c}
\hline \thickhline

 \rowcolor[HTML]{f8f9fa} \multicolumn{1}{c||}{Input Clip Length $s$} & \multicolumn{1}{c}{UCF-Crime} \\
\rowcolor[HTML]{f8f9fa} \multicolumn{1}{c||}{(Number of frames)} & \multicolumn{1}{c}{(AUC\%)}  \\\hline
\hline
1 & 84.25  \\
3 & 84.56  \\
5 & \textbf{84.89}  \\
7 & 83.15 \\
\hline
\end{tabular}\hspace{9mm}
}

\subfloat[Analysis of Inference speed. \label{tab:Analysis of Inference speed}]
{
   \renewcommand{\arraystretch}{1.1}
    \setlength{\tabcolsep}{5pt}
   \begin{tabular}{c||c}
\hline \thickhline

 \rowcolor[HTML]{f8f9fa} \multicolumn{1}{c||}{Datasets} & \multicolumn{1}{c}{Average speed of inference} \\
\rowcolor[HTML]{f8f9fa} \multicolumn{1}{c||}{Name} & \multicolumn{1}{c}{(FPS)}  \\\hline
\hline
UCF-Crime & 0.82   \\
XD-Violence & 0.86   \\
UBnormal & 0.79  \\
CSAD & 0.53  \\
\hline
\end{tabular}\hspace{9mm}
}

}
\label{tab:addition-ablation-multi}
\end{table}

\subsection{Effect of Input Clip Length}

Table~\ref{tab:Effect of input clip length} analyzes how varying the number of frames in each input video clip affects PANDA’s performance. As the input length increases from 1 to 5 frames, detection accuracy steadily improves, suggesting that short-range temporal cues are beneficial to the reasoning process. However, when the clip length is extended to 7 frames, performance noticeably drops. We hypothesize that this is due to the binary labeling strategy used during evaluation—if a clip is anomalous, all frames of the clip are scored as anomalous. For longer clips that may contain both normal and abnormal frames, this scoring scheme introduces noise, leading to performance degradation.

\subsection{Analysis of Inference Speed}

Table~\ref{tab:Analysis of Inference speed} reports the average inference speed of PANDA across different datasets. As observed, PANDA achieves similar inference times on UCF-Crime, XD-Violence, and UBnormal. However, a noticeable slowdown is observed on the CSAD dataset. This is primarily because CSAD contains videos with complex conditions and scenarios, leading PANDA to more frequently enter the reflection stage. Since the reflection stage involves invoking additional tools, it introduces greater computational overhead. Despite this, the overall average inference speed of PANDA remains acceptable for non-time-sensitive applications, demonstrating its practical feasibility in real-world deployments where latency is not a critical constraint.

\section{Toolset Details}
\label{Toolset}

PANDA is equipped with a modular and extensible set of tools designed to enhance video content analysis, mitigate visual degradation, and provide external contextual information. These tools are automatically selected and invoked during the self-reflection stage based on the detection context. Below, we provide a detailed summary of each tool integrated into the PANDA framework.
\vspace{-2mm}
\paragraph{Object Detection.} PANDA employs the YOLOWorld~\cite{cheng2024yolo} model pretrained on a wide set of open-world concepts. It supports fine-grained category-specific detection including actions like ``person hitting another'', ``person setting something on fire'', and ``person stealing''. This enables robust anomaly-related scene understanding through bounding-box localization and category labels.
\vspace{-2mm}
\paragraph{Image Denoising.} PANDA uses OpenCV’s fast non-local means filter for image denoising. It reduces color and spatial noise in frames using adaptive filtering, helping enhance clarity in low-light or noisy environments.
\vspace{-2mm}
\paragraph{Image Deblurring.} PANDA applies unsharp masking with Gaussian blur subtraction to sharpen edge details for motion blur or out-of-focus issues. This lightweight enhancement improves perceptual clarity without the need for retraining.
\vspace{-2mm}
\paragraph{Image Brightness Enhancement.} PANDA uses OpenCV’s CLAHE (Contrast Limited Adaptive Histogram Equalization) on the L-channel in LAB color space. This ensures localized brightness normalization for dimly lit or overexposed frames.
\vspace{-2mm}
\paragraph{Image Super-Resolution.} PANDA integrates the Real-ESRGAN~\cite{wang2021real} model for resolution enhancement. It improves detail preservation and restores textures in low-resolution videos using a deep RRDBNet-based super-resolution pipeline.
\vspace{-2mm}
\paragraph{Image Retrieval.} PANDA uses CLIP-based~\cite{radford2021learning} visual-textual retrieval to match current queries (e.g., ``robbery incident'') with previously seen keyframes. Cosine similarity between CLIP embeddings is used for scoring relevance.
\vspace{-2mm}
\paragraph{Web Search.} PANDA leverages the Tavily Search API~\cite{tavily} for querying web content related to unknown or uncertain anomalies. Search results are parsed into structured summaries that can be referenced in the reasoning process.
\vspace{-2mm}
\paragraph{Image Zooming.} PANDA leverages a bicubic interpolation-based zooming tool to magnify regions that require enhanced spatial detail using a specified zoom factor. This is useful when detecting small-scale interactions or distant activities.

All tools are dynamically invoked during the reflection stage via the MLLM-generated reflection plan. Each tool outputs enhanced frame sets and structured summaries that are used to augment the reasoning prompt for follow-up reasoning steps.

\section{System Prompts}
\label{System Prompt}
In this section, we present the detailed prompts used by PANDA across its core stages. Figure \ref{fig:knowledge base constructing prompt} illustrates the prompt used during anomaly knowledge base construction. Figure \ref{fig:perception prompt} shows the self-adaptive environmental perception prompt. Figure \ref{fig:Planning prompt} presents the prompt for anomaly detection strategy planning. Figure \ref{fig:Reasoning prompt} demonstrates the goal-driven heuristic reasoning prompt. Figure \ref{fig:Reflection prompt} displays the prompt used during the tool-augmented self-reflection phase.

\section{Additional Visualization Results}
\label{Additional visualization results}
Figures \ref{fig:Vis_XD} and \ref{fig:Vis_UB} further show the qualitative results of the samples on the XD-Violence and UBnormal test sets.

\section{Discussions}
\label{Discussions}

\subsection{Limitations}
PANDA currently integrates a curated set of commonly used tools for enhancement and reasoning. While these tools suffice for most general scenarios, expanding the toolkit to accommodate domain-specific modalities (e.g., thermal imaging) could broaden PANDA’s applicability.

\subsection{Broader Impacts}

PANDA advances the paradigm of automated video anomaly detection by integrating vision-language models and decision-time tool augmentation into a unified AI agent framework. This work has the potential to improve the robustness and interpretability of security monitoring systems, enabling adaptive deployment across diverse environments without training. Moreover, We also recognize the importance of ethical considerations: systems like PANDA must be deployed responsibly, with attention to privacy protection, fairness, and minimizing unintended surveillance harms.

\subsection{Future Work}

\textbf{Enhancing Real-Time Adaptability.} \quad To make PANDA more suitable for practical deployments, future work could focus on reducing inference latency and optimizing tool invocation paths. 

\textbf{Improving Spatial Reasoning and Localization.} \quad Currently, PANDA focuses on frame-level or clip-level anomaly identification. Incorporating spatial anomaly localization, such as identifying the precise region or object involved in an abnormal event, could significantly expand its utility in surveillance systems.

\begin{figure*}[t]
  \centering
   \includegraphics[width=1\linewidth]{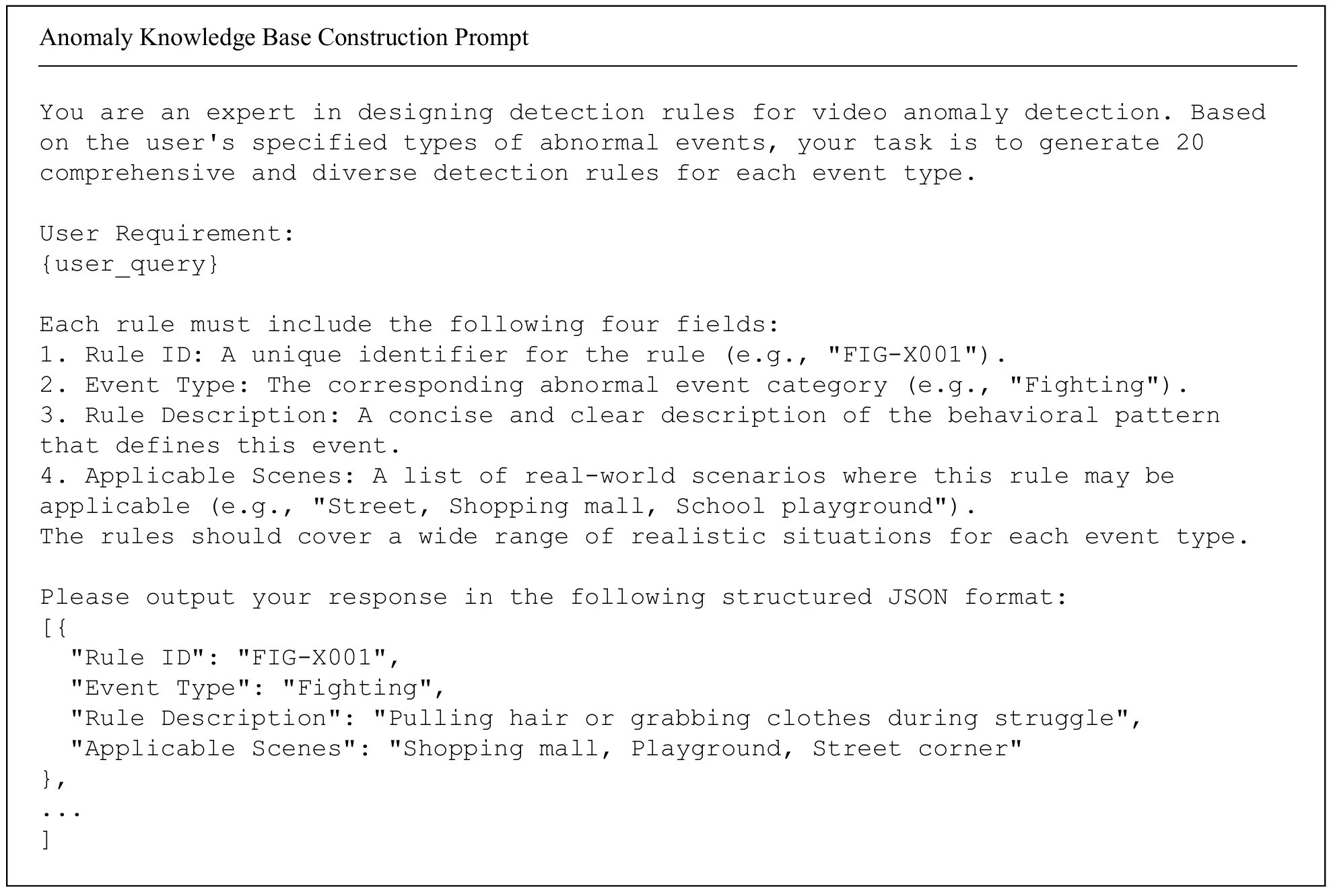}
    \caption{System prompt for anomaly knowledge base construction.}
   \vspace{-3mm}
   \label{fig:knowledge base constructing prompt}
\end{figure*}

\begin{figure*}[t]
  \centering
   \includegraphics[width=1\linewidth]{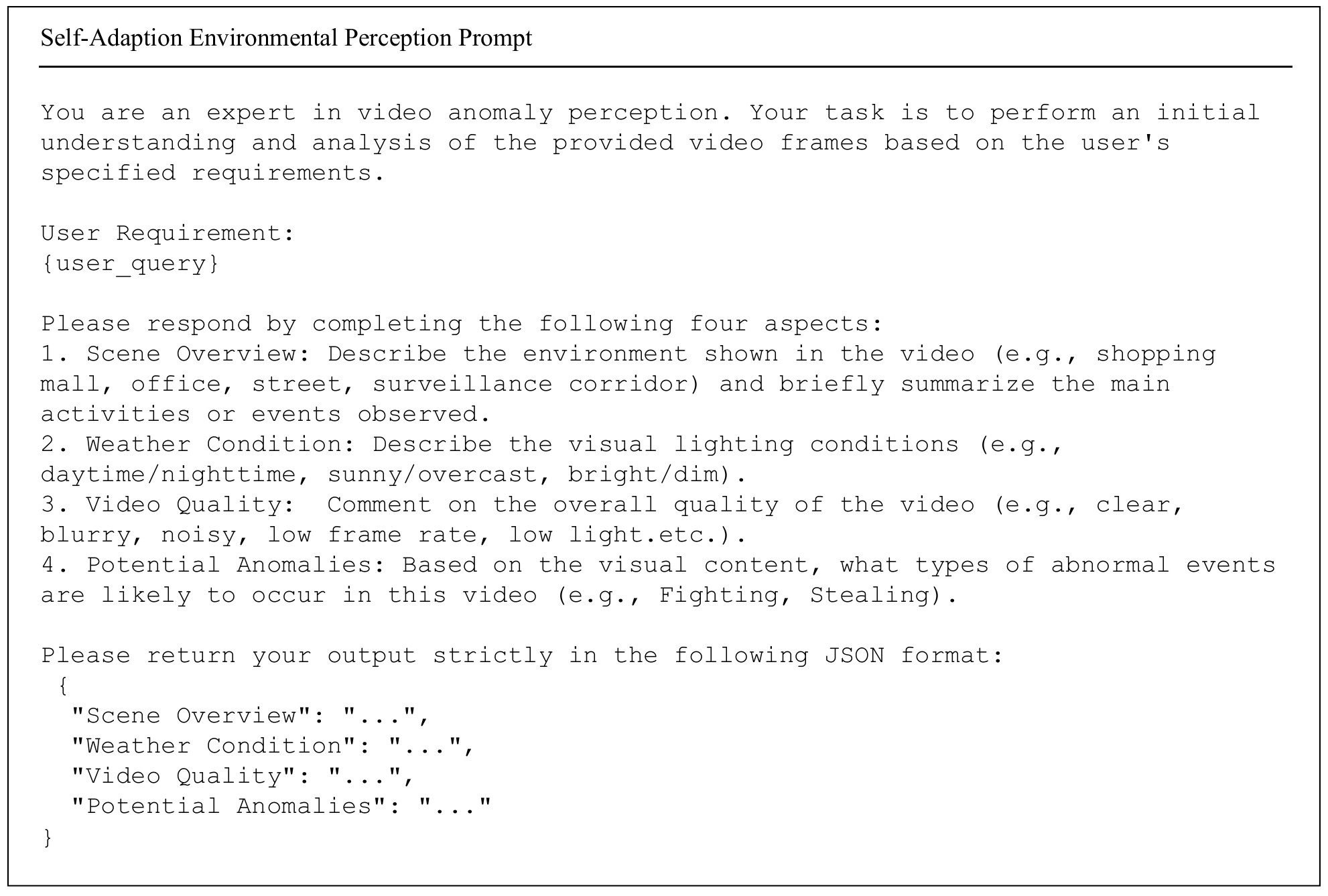}
    \caption{System prompt for self-adaptive environmental perception.}
   \vspace{-3mm}
   \label{fig:perception prompt}
\end{figure*}

\begin{figure*}[t]
  \centering
   \includegraphics[width=1\linewidth]{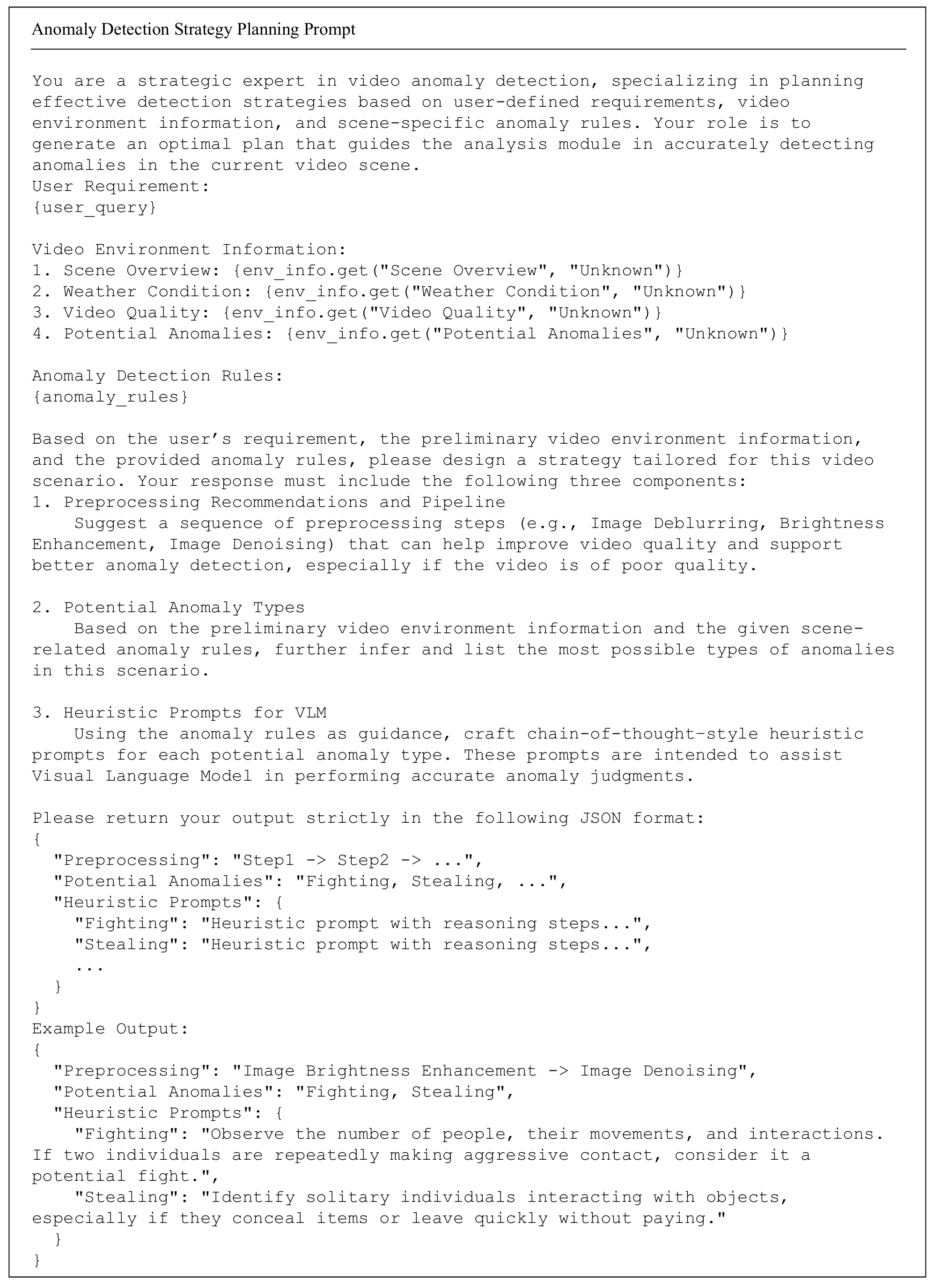}
    \caption{System prompt for anomaly detection strategy planning.}
   \vspace{-3mm}
   \label{fig:Planning prompt}
\end{figure*}

\begin{figure*}[t]
  \centering
   \includegraphics[width=1\linewidth]{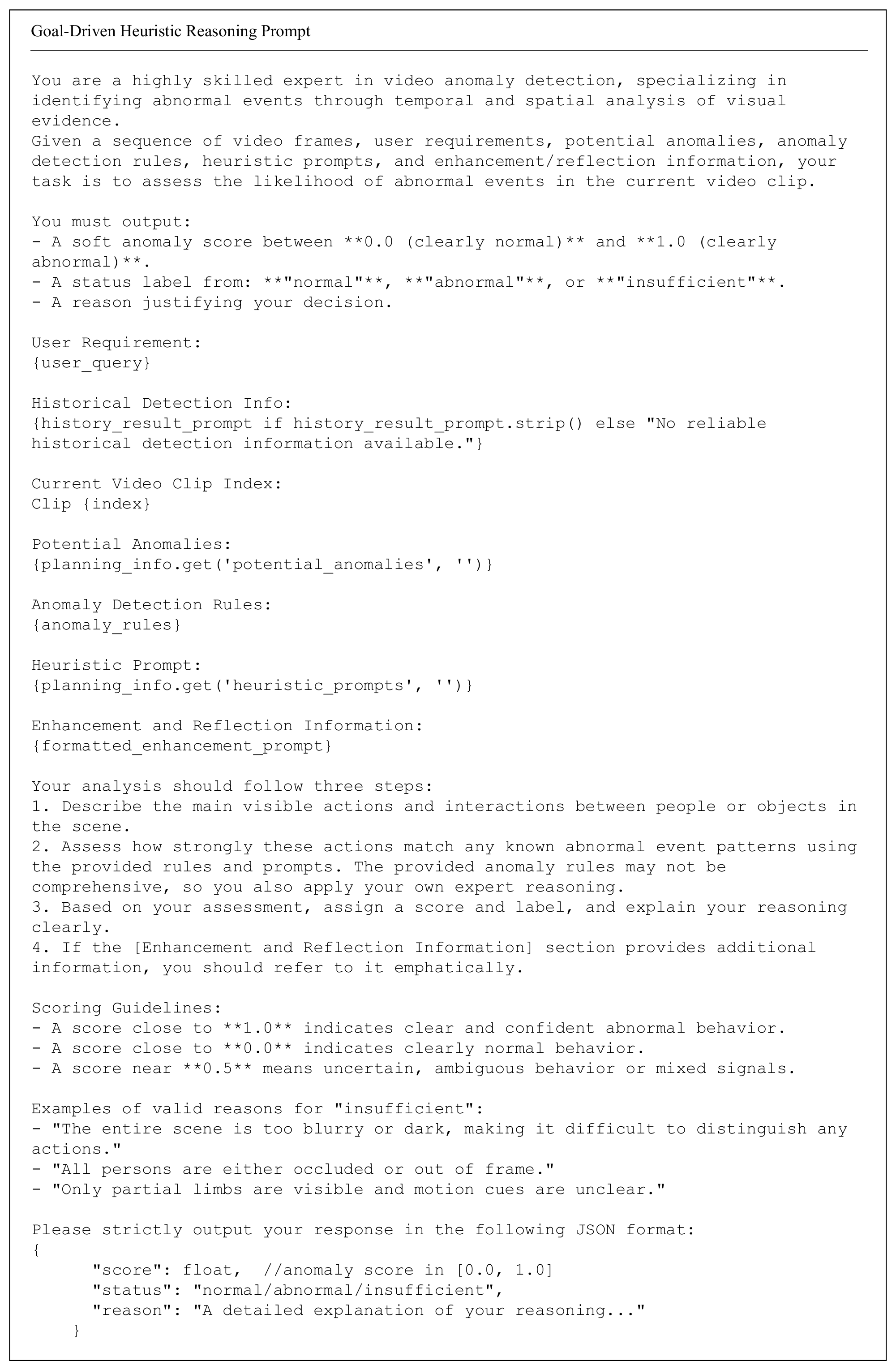}
    \caption{System prompt for goal-driven heuristic reasoning.}
   \vspace{-3mm}
   \label{fig:Reasoning prompt}
\end{figure*}

\begin{figure*}[t]
  \centering
   \includegraphics[width=1\linewidth]{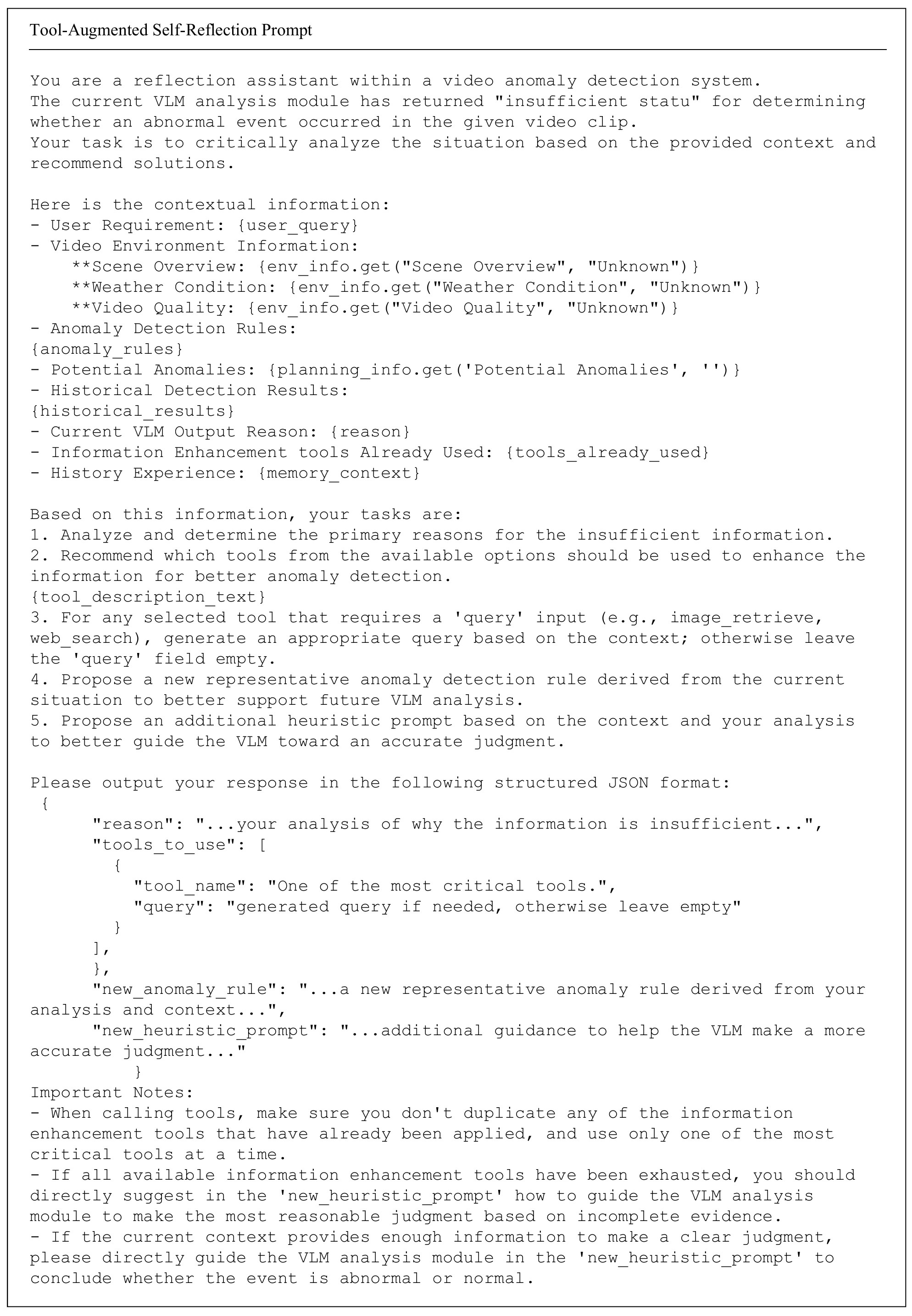}
    \caption{System prompt for tool-augmented self-reflection.}
   \vspace{-3mm}
   \label{fig:Reflection prompt}
\end{figure*}

\begin{figure*}[t]
  \centering
   \includegraphics[width=1\linewidth]{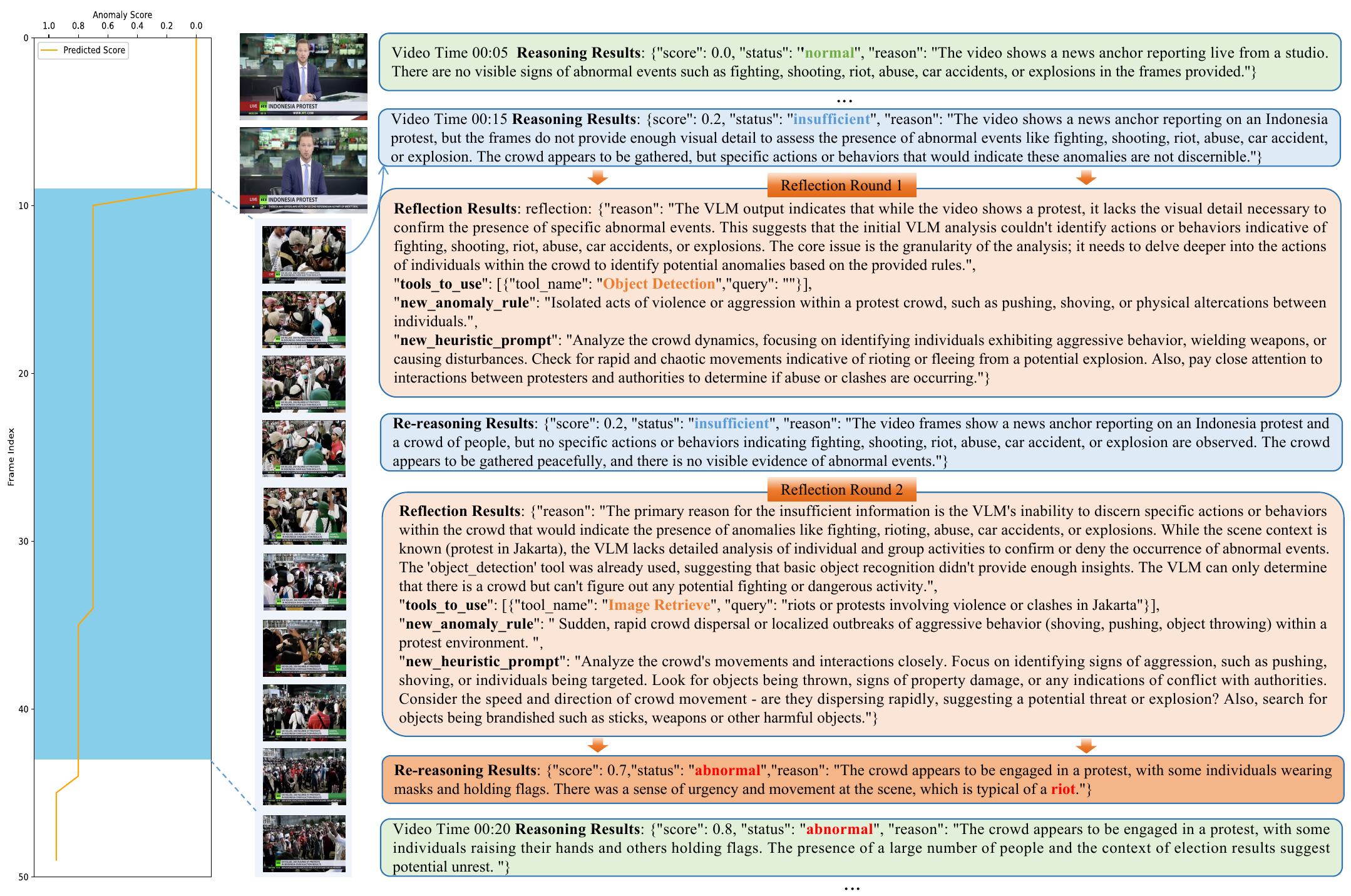}
    \caption{Visualization of qualitative results for a sample on the XD-Violence test set.}
   \vspace{-3mm}
   \label{fig:Vis_XD}
\end{figure*}

\begin{figure*}[t]
  \centering
   \includegraphics[width=1\linewidth]{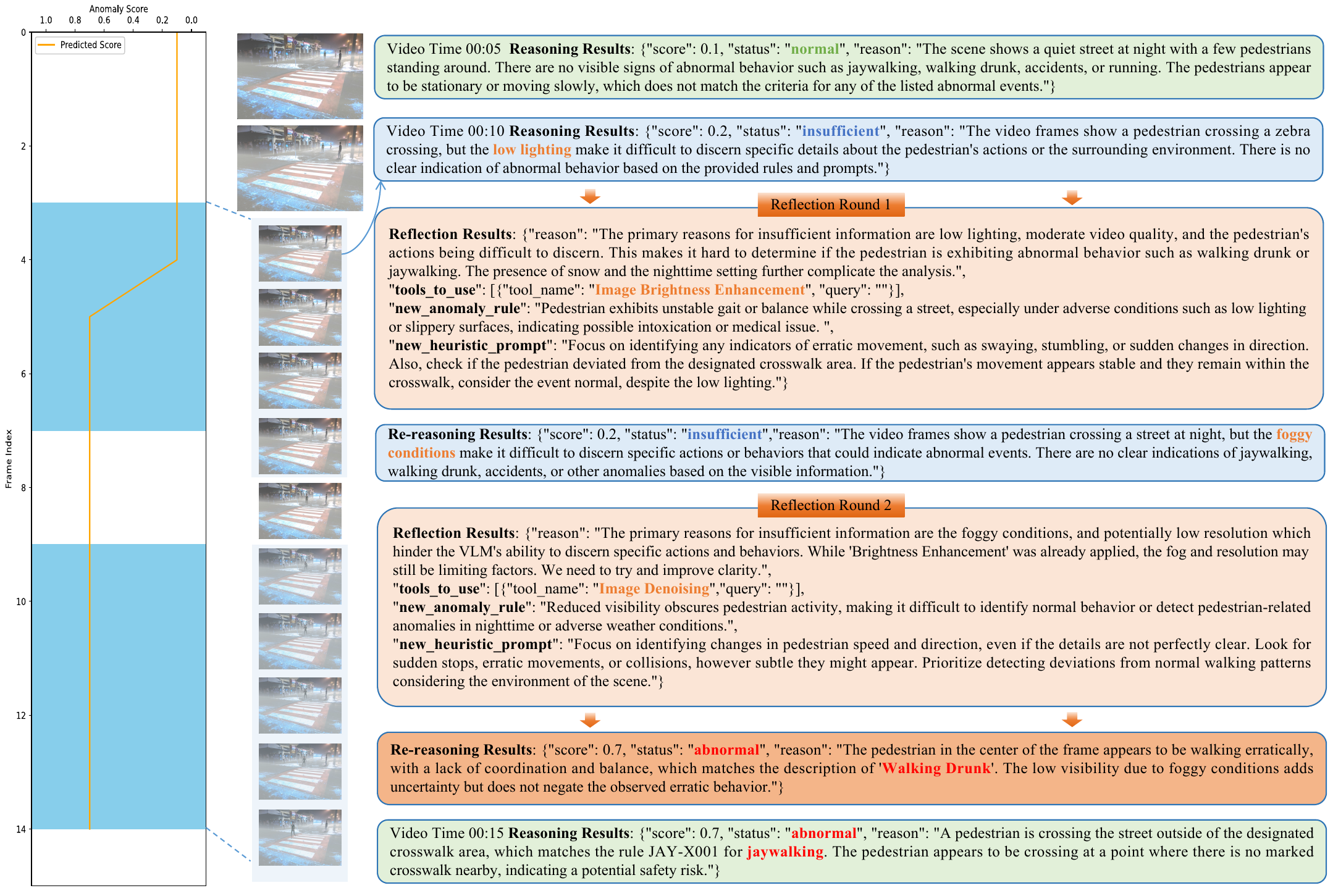}
    \caption{Visualization of qualitative results for a sample on the UBnormal test set.}
   \vspace{-3mm}
   \label{fig:Vis_UB}
\end{figure*}





}

\end{document}